\documentclass[10pt,twocolumn,letterpaper]{article}

\usepackage{iccv}
\usepackage{times}
\usepackage{epsfig}
\usepackage{graphicx}
\usepackage{amsmath}
\usepackage{amssymb}

\usepackage[caption=false,font=footnotesize]{subfig}

\usepackage[breaklinks=true,bookmarks=false]{hyperref}

\iccvfinalcopy 

\def\iccvPaperID{****} 

\ificcvfinal\fi

\usepackage{algorithm}
\usepackage{algorithmic}
\usepackage{epsfig}
\usepackage{graphicx}
\usepackage{amsmath}
\usepackage{amssymb}
\usepackage{url}            
\usepackage{booktabs}       
\usepackage{amsfonts}       
\usepackage{nicefrac}       
\usepackage{microtype}      
\usepackage{mathtools}
\usepackage{xspace}
\usepackage{bbm}
\usepackage{color}
\usepackage{multirow}
\usepackage[group-separator={,}, group-minimum-digits={3}]{siunitx}
\usepackage{textcomp}
\usepackage{gensymb}
\usepackage{caption}
\usepackage{tabularx}
\usepackage{multirow}
\usepackage{makecell}
\makeatletter
\DeclareRobustCommand\onedot{\futurelet\@let@token\@onedot}
\def\@onedot{\ifx\@let@token.\else.\null\fi\xspace}

\def\eg{\emph{e.g}\onedot} 
\def\ie{\emph{i.e}\onedot}

\def\etal{\emph{et al}\onedot}
\makeatother

\usepackage{color}
\usepackage{bm}
\usepackage{amsfonts,amssymb}
\usepackage{bbm}

\usepackage[T1]{fontenc}
\usepackage[utf8]{inputenc}
\usepackage{authblk}

\usepackage{fancyhdr} 

\begin{document}

\title{You Cannot Easily Catch Me: A Low-Detectable Adversarial Patch for Object Detectors}


\author[1]{Zijian Zhu}
\author[2]{Hang Su}
\author[1]{Chang Liu}
\author[1]{Wenzhao Xiang}
\author[1]{Shibao Zheng}
\affil[1]{Institute of Image Communication and Networks Engineering
in the Department of Electronic Engineering~(EE), Shanghai Jiao Tong University}
\affil[2]{Department of Computer Science and Technology, Institute for AI, THBI Lab, Tsinghua University}
 
\renewcommand\Authands{ and }

\maketitle
\ificcvfinal\thispagestyle{empty}\fi

\begin{abstract}
Blind spots or outright deceit can bedevil and deceive machine learning models. Unidentified objects such as digital ``stickers,'' also known as adversarial patches, can fool facial recognition systems, surveillance systems and self-driving cars. Fortunately, most existing adversarial patches can be outwitted, disabled and rejected by a simple classification network called an adversarial patch detector, which distinguishes adversarial patches from original images. An object detector classifies and predicts the types of objects within an image, such as by distinguishing a motorcyclist from the motorcycle, while also localizing each object's placement within the image by “drawing” so-called bounding boxes around each object, once again separating the motorcyclist from the motorcycle. To train detectors even better, however, we need to keep subjecting them to confusing or deceitful adversarial patches as we probe for the models’ blind spots. For such probes, we came up with a novel approach, a Low-Detectable Adversarial Patch, which attacks an object detector with small and texture-consistent adversarial patches, making these adversaries less likely to be recognized. Concretely, we use several geometric primitives to model the shapes and positions of the patches. To enhance our attack performance, we also assign different weights to the bounding boxes in terms of loss function. 
Our experiments on the common detection dataset COCO as well as the driving-video dataset $\text{D}^2$-City show that LDAP is an effective attack method, and can resist the adversarial patch detector.
\end{abstract}

\section{Introduction}
Computer vision for autonomous driving systems remains highly vulnerable to failures of object detection. Despite impressive recent improvements for locating and classifying objects in digital images  ~\cite{ren2015faster,liu2016ssd,he2017mask,redmon2018yolov3}, object detectors generally remain quite vulnerable to patch-wise adversarial threats ~\cite{liu2019dpatch,thys2019fooling,wu2020making,huang2020universal}. Deadly traffic accidents might result from attacks on detection-based autonomous driving systems. For example, an autonomous vehicle might hit a pedestrian not properly detected due to an adversarial attack.

Fortunately, most existing patch-wise adversarial examples can be distinguished readily from  benign images, making it easy to train a classification network~(called an \textbf{adversarial patch detector}) to automatically detect and reject these adversarial images.
This simple defense method, also known as a detect-only defense approach~\cite{akhtar2018threat}, can disable most existing patch-wise adversarial attacks.
We refer to this drawback of existing patch-wise adversarial attacks as \textbf{high detectability}.
Although recognition of this drawback can help researchers easily add features to protect detection models from attacks, this approach also prevents researchers from revealing more potential security problems with these detection models.
To overcome this drawback, it is important to study patch-wise attacks that can escape notice by an adversarial patch detector.

One possible way to design such an attack is to make adversarial examples indistinguishable from benign images, making it harder for the adversarial patch detector to recognize adversarial examples. 
To achieve this goal, we focus on two aspects of the attack method, including reducing the area of adversarial patches in the images, and increasing the texture consistency with the original images.
Both aspects can reduce the difference between adversarial examples and original images, making adversarial examples less likely to be recognized by an adversarial patch detector.

\begin{figure}[t]
\centering
\subfloat[]{\includegraphics[width=0.3\linewidth]{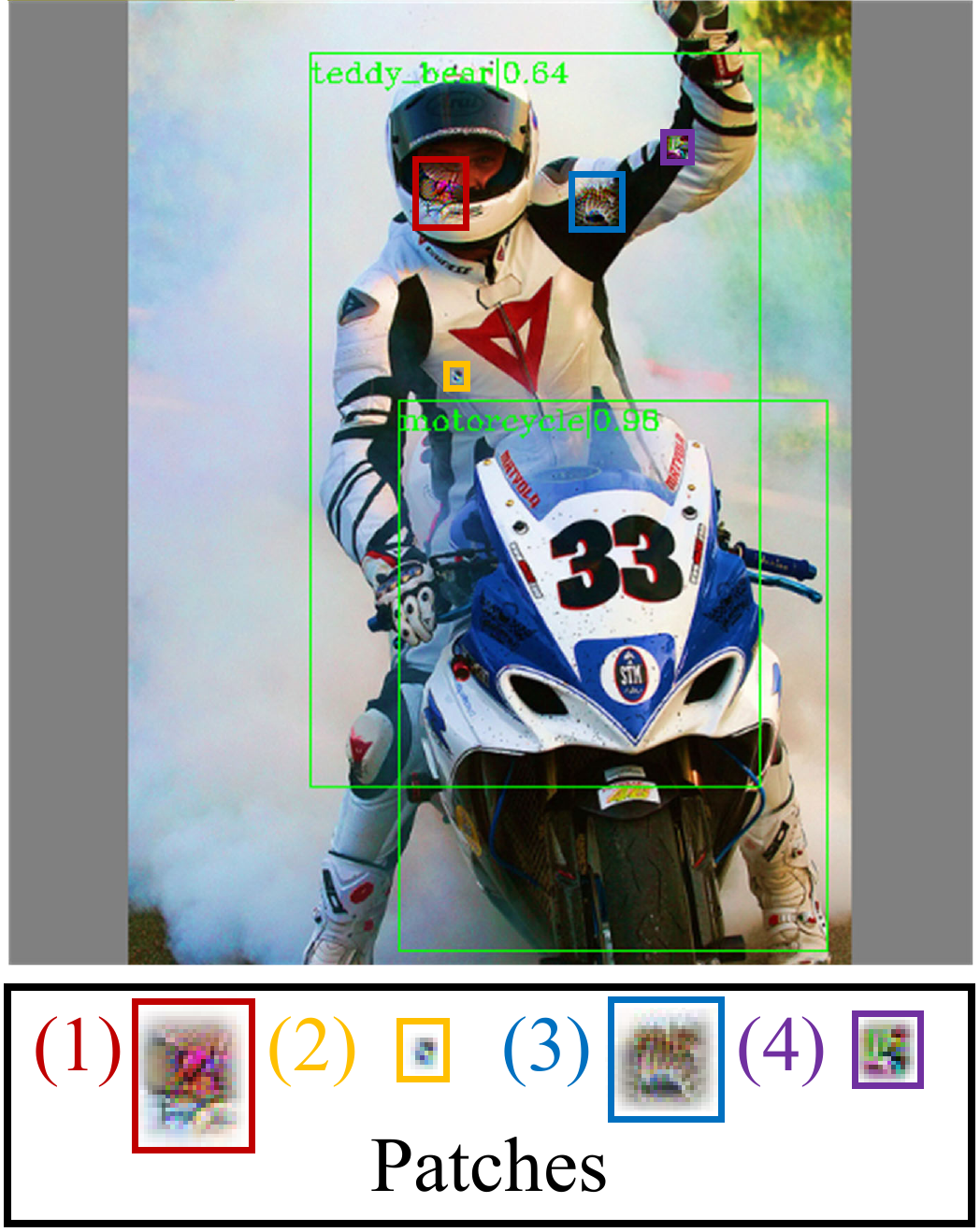} 
}
\subfloat[]{\includegraphics[width=0.3\linewidth]{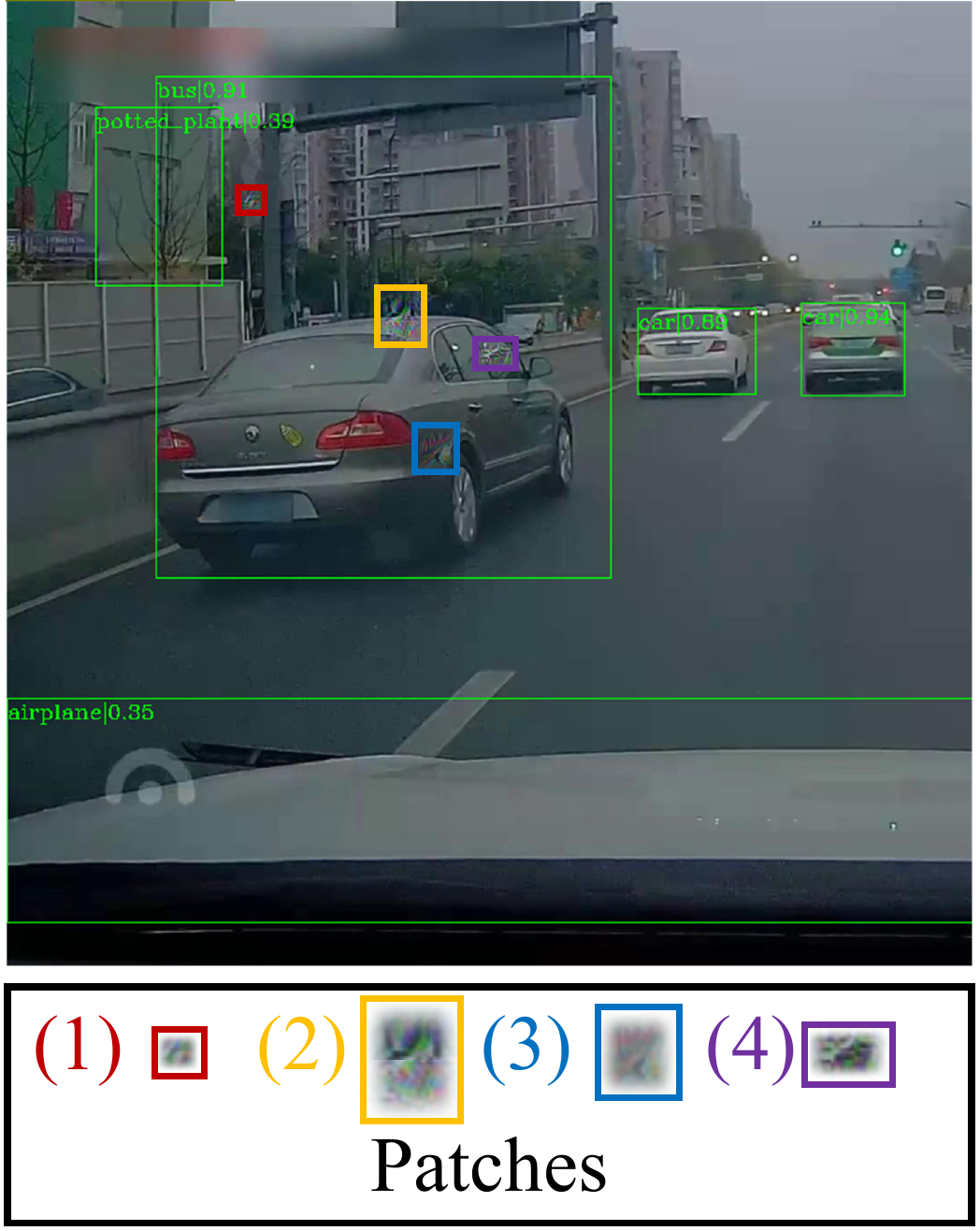} 
}
\subfloat[]{\includegraphics[width=0.3\linewidth]{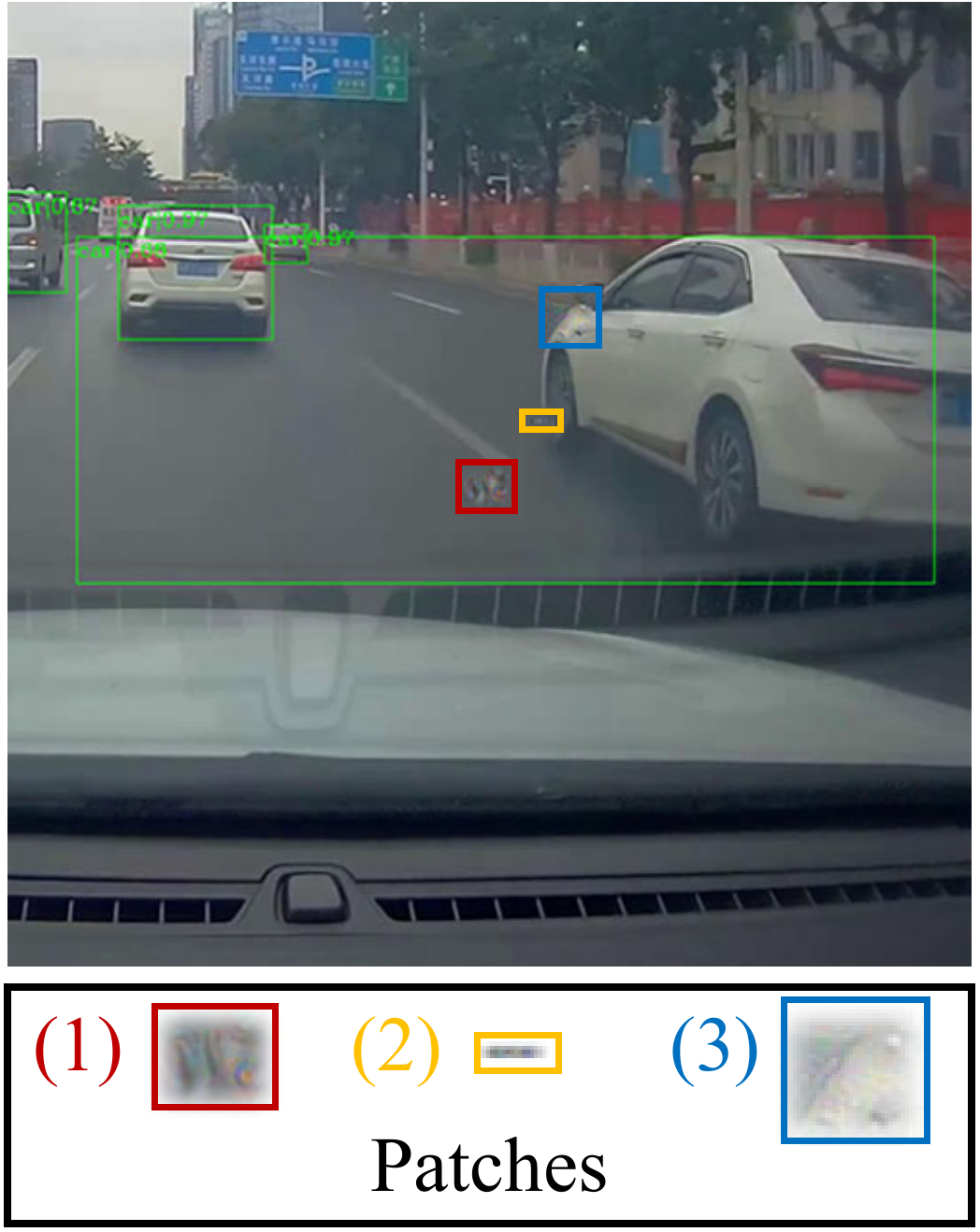} 
}
\caption{ \small
Examples of LDAP attacking Faster R-CNN~\cite{ren2015faster}.
(a):~classification attack on COCO; the person is detected as a teddy bear with 64\% confidence. 
(b):~classification attack on $\text{D}^2$-City; the car is detected as a bus with 91\% confidence. 
(c):~localization attack on $\text{D}^2$-City; the predicted bounding box only has 40.63\% IoU with the ground truth bounding box of the target object.
We also zoom in on the patches and show each one of them below the images.
We find that all these patches are small and maintain consistent texture with the original images.
}
\label{fig:demo_attack}
\vspace{-1em}
\end{figure}

\subsection{Challenges}
However, it is non-trivial to deceive an object detector while simultaneously reducing the area of adversarial patches, due to the following reasons.

\subsubsection{Model the  Patch Shapes Parametrically}
It is non-trivial to model patch shapes parametrically, since their shapes can be flexible.
One simple method is to model all patches in an input image as a binary mask covering the image, and exhaust any possible shapes of patches with respect to every pixel in it. However, this modeling makes the problem difficult to solve, for it is a high-dimensional discrete-search problem.

\subsubsection{The Inner Mechanism of an Object Detector}
Assuming that we are able to model the patches, it is still difficult to attack an object detector, especially with flexible patches. 
The difficulty mainly comes from the bounding-box prediction mechanism of an object detector. Generally the pipeline of an object detector is: firstly it predicts redundant bounding boxes on one object, then the detector classifies them, and lastly it uses Non-Maximum Suppression~(NMS)\footnote{NMS is a technique to select one bounding box from multiple bounding boxes based on their categories.} to select one bounding box from them~\cite{ren2015faster,liu2016ssd,he2017mask,redmon2018yolov3}. The attack should simultaneously influence all of these bounding boxes, which makes the attack on object detectors difficult.
To address this issue, previous attacks with fixed patches adopted two strategies.
The first was to attack the bounding box with a maximum confidence score for the target category~\cite{liu2019dpatch,eykholt2018physical,thys2019fooling}. However, during optimization the bounding box with maximum category score~(\ie, the output of the model ought to represent the confidence of classification) can be continuously changed, rendering the search-for-region unstable in our method.
The second strategy was to equally attack the bounding boxes that can be detected as the target object, which is the object that the adversary aims to attack~\cite{huang2020universal}. Nonetheless, most of these bounding boxes are inaccurate, misleading the region to sub-optimal positions. 
Therefore, neither of these strategies is suitable for our region-flexible attack.

\subsection{Our Proposal}

To address the aforementioned challenges, we propose a Low Detectable Adversarial Patch~(\textbf{LDAP}).
We formulate the generation of low-detectable patches as a joint optimization problem that corrupts the prediction of the object detector; reduces the area of patches; and simultaneously preserves the texture consistency with the original image.
To achieve this, we solve the challenges from the following two perspectives.

\subsubsection{The Use of Geometric Primitives to Model the Patches}

Some previous research represents a given two-dimensional~(2-D) shape with a set of geometric primitives~(ellipses)~\cite{da2010fitting}\cite{panagiotakis2016parameter}. 
They utilize both a Gaussian Mixture Model~(GMM) to parameterize these ellipses as well as an Expectation-Maximization~(EM) algorithm to maximize the similarity to the given shape.
Referring to this work, we use a set of geometric primitives~(rectangles) to model the patches with flexible shapes and positions.
Concretely at first, based on a finding that patches in irregular shapes on an image plane can be approximated by several rectangles, we assume that the adversarial patches can be approximated by a set of rectangles. 
Therefore, we model the region of patches as a union of several rectangles. The division of these patches from other regions can be realized by a binary image mask, in which the patch regions are set to 1 and the others are set to 0.
As a result, the optimization of shapes and positions of patches is converted to the optimization of shape and position parameters of these rectangles.

However, the boundaries of these rectangles are steep, making the loss function not differentiable with respect to their shape and position parameters. To address this issue, we relax the steep boundaries of these rectangles to smooth transitions from 1 to 0.
These boundary-relaxed rectangles become continuous everywhere, enabling us to optimize their parameters using a gradient-based method. 
The relaxation makes the rectangles look feathered, as shown in the attack examples of LDAP in Fig.~\ref{fig:demo_attack}.

\subsubsection{Soft-Attack Strategy}
To overcome the difficulty from the inner mechanism of an object detector, we adopt a soft-attack strategy in the loss function, meaning that we assign different weights to multiple bounding boxes that can be detected as a target object. 
This strategy is based on the assumption that the attack should pay more attention to the bounding boxes that are closer to the target object. 
This assumption is reasonable, since the closer bounding boxes are more likely to be the correct predictions.
The weights are set as the Intersection over Union~(IoU) between the predicted bounding box and the bounding box of the target object; IoU is a widely-used metric in detection literature for measuring the overlap between bounding boxes.
The soft-attack strategy in loss function enables our attack to be more focused on the bounding boxes that are closer to the target object, a focus that benefits our attack when searching for patch positions.

We evaluate the effectiveness of LDAP on the common detection dataset COCO, and validate the threat of LDAP to an autonomous driving application on the driving-video dataset $\text{D}^2$-City.
We test LDAP on four mainstream detection models, including Faster R-CNN~(FRCN)~\cite{ren2015faster}; Mask R-CNN~(MRCN)~\cite{he2017mask}; SSD~\cite{liu2016ssd}; and YOLOv3~\cite{redmon2018yolov3}.
As shown in Fig.~\ref{fig:demo_attack}, LDAP can deceive the detector with small and texture-consistent patches in these images.

We conclude our main contributions as follows:
\begin{itemize}

    \item We propose a low-detectable patch-wise attack method. Based on a reasonable assumption, 
    we formulate the low-detectable attack as a joint optimization problem that simultaneously takes into consideration the attack performance, patch area and texture consistency of patches;
    
    \item We model the shapes and positions of patches in a differentiable way, which enables us to use a gradient-based method to optimize shapes and positions of patches together with their textures; 
    
    \item To overcome the difficulty from the inner mechanism of an object detector, we adopt a soft-attack strategy in our loss function, assigning different weights to bounding boxes that can be detected as the target object.

\end{itemize}

The remainder of the paper is organized as follows: In Section~\ref{sec:Related_Works}, we introduce previous research related to our methods and, in Section~\ref{sec:Methodology},
we propose a low-detectable adversarial patch;  our experimental results are presented in Section~\ref{sec:Experiments}, followed by a conclusion.

\section{Related Work}
\label{sec:Related_Works}
In this section, we review three related bodies of work in terms of object detection, adversarial attack and 2-D shape modeling.

\subsection{Object Detection}
First, we introduce object-detection models, which are the target models we aim to attack.
Nowadays, mainstream object detectors can be classified into two-stage detectors and one-stage detectors. 

\subsubsection{Two-stage Detection}
The prediction procedure of two-stage detectors consists of a region proposal followed by a classification. 
Detectors of this type first search for region proposals, then classify each of them. 
An early detector with a Convolution Neural Network~(CNN) is Overfeat~\cite{sermanet2014overfeat}, which combines a sliding window with a CNN to do the detection. After that, a so-called Regions with CNN features~(R-CNN) detector is proposed~\cite{girshick2014rich}. R-CNN uses selective search to get proposals, and classifies them based on CNN features.
After R-CNN, a number of studies were published to speed up the detection process and improve accuracy. Representative modern detectors include Fast R-CNN~\cite{girshick2015fast}, Faster R-CNN~\cite{ren2015faster} and Mask R-CNN~\cite{he2017mask}.

\subsubsection{One-stage Detection}
One-stage detectors, also known as ``single-shot detectors,'' simultaneously predict both the bounding boxes and the category scores with just a single pass through the network. The representative one-stage detectors are YOLO series~\cite{redmon2016you}\cite{redmon2017yolo9000}\cite{redmon2018yolov3}\cite{bochkovskiy2020yolov4} and SSD~\cite{Liu_2016ssd}. 
Compared with two-stage detectors, one-stage detectors run faster, but usually get lower detection accuracy. 
In our experiments, we find that we need smaller patches to attack one-stage detectors than two-stage detectors, a finding that indicates that one-stage detectors are easier to attack.

\subsection{Adversarial Attacks}
Adversarial attacks on computer vision systems have been widely studied. In the research field of object-detection attack, many attack algorithms have been proposed in recent years. From the perspective of manipulated region, we classify these adversarial attacks into pixel-wise attacks and patch-wise attacks.

\subsubsection{Pixel-wise Attacks}
Pixel-wise attacks change all the pixels of the target object or the whole image slightly; consequently, only rarely can human beings perceive such perturbations.
This kind of attack is firstly studied in classification attacks, \eg, FGSM~\cite{goodfellow2014explaining}, DeepFool~\cite{moosavi2016deepfool} and PGD~\cite{madry2018towards}, and then it is extended to object detection attack.
Xie~\etal~\cite{xie2017adversarial} propose Dense Adversary Generation~(DAG), which is an attack method for detection and segmentation. In detection attack, DAG changes pixels of the whole image to increase the wrong-class confidence and decrease correct-class confidence. 
Wei~\etal~\cite{wei2018transferable} propose an attack scheme based on a conditional Generative Adversarial Network to generate an adversarial perturbation that changes the pixels in the whole image to deceive the detector. 
Nezami~\etal~\cite{nezami2021pick} propose an attack method that manipulates the pixels of the target object and precisely changes the label of the target object, without changing the labels of other objects.

\subsubsection{Patch-wise Attacks}
These attacks try to deceive the detector by manipulating small regions in the input image, called patches. The positions of the patches are usually predefined; for example, the upper-left corner of the input image or the center of the target object.
DPatch~\cite{liu2019dpatch} is an early work of a patch-wise object detection attack. It attaches an adversarial patch on the upper-left corner of an image, and is able to disturb the digital object-detection.
Some methods that emerged later are capable of being transferred from the digital world to the physical. 
Eykholt~\etal~\cite{eykholt2018physical} extend the Robust Physical Perturbations~($RP_2$) algorithm~\cite{eykholt2018robust} from classification attack to object-detection attack. The sticker version of their extended $RP_2$ algorithm, denoted as $RP_2$-Sticker, puts two adversarial patches on a stop sign, and is able to mislead the detector to predict the stop sign as other objects.
Thys~\etal~\cite{thys2019fooling} train their adversarial pattern, denoted as AdvPatch, and print it on a paper board. The volunteers holding this paper board successfully evade the detection of YOLOv2~\cite{redmon2017yolo9000}.
Meanwhile, Xu~\etal~\cite{xu2020adversarial} print their adversarial patterns on the front of a T-shirt while considering the non-rigid deformation of the pattern caused as the wearer's pose changes. Volunteers wearing this T-shirt are neglected by the detector.
Huang~\etal~\cite{huang2020universal} 
propose Universal Physical Camouflage~(UPC), which can mislead detectors to predict the volunteers as dogs when they wear clothes with specially designed camouflages.

However, the adversarial patches of these patch-wise attacks are all large and have inconsistent textures compared with the original image, making them easy to be caught by an adversarial patch detector and then rejected. This defense method can disable most existing patch-wise attack methods. We regard this drawback as high detectability. In our method, we try to establish a low-detectable patch-wise digital attack, to escape the adversarial patch detector.

\subsection{2-D-Shape Modeling}
One aspect of our method involves modeling the shapes of patches. There exists some research about how to model a 2-D shape. 
Da and Kemp~\cite{da2010fitting} use multiple geometric primitives~(ellipses) to fit a given human upper-body silhouette, which is a complex shape, then model the shape with GMM, and utilize unconstrained Expectation-Maximization~(EM) to optimize the parameters of GMM to fit the given shape.
Panagiotakis and Argyros~\cite{panagiotakis2016parameter} design a  shape-modeling method, also using geometric primitives~(ellipses) to represent the given shape.
Modeling the ellipses with GMM, they use hard EM and Akaike Information Criterion to optimize the parameters of GMM to approximate the given shape.
We borrow their modeling idea, and assume that a complex shape, even several shapes, can be modeled as the combination of several geometric primitives. We introduce how we model the shapes of patches in detail in Section~\ref{sec:Area_Loss}.

\section{Methodology}
\label{sec:Methodology}
It is complicated to generate a low-detectable adversarial patch. In Section~\ref{sec:Problem_Formulation}, we first formulate the problem in such a way that we reduce the combination of three loss functions. Then we introduce each loss function in Sections~\ref{sec:Attack_Loss} to~\ref{sec:Texture_Loss}. Finally, we introduce the optimization method in Section~\ref{sec:Optimization}.

\subsection{Problem Formulation}
\label{sec:Problem_Formulation}
The goal of our LDAP is to deceive the object detector with small and texture-consistent patches.
We set this goal based on an assumption that both aspects can help us evade an adversarial patch detector.
The assumption is reasonable, since small patches can reduce the area of the manipulated part in the image, and consistent texture in these patches can reduce the magnitude of manipulation. Both aspects help us decrease the difference between our adversarial example and the benign image, making it less likely for an adversarial patch detector to distinguish them.

Towards this goal, given an input image $I\in R^{W\times H}$\footnote{For clarity, we omit the 3-channels~(RGB) for images and perturbations in all formulae, which does not influence the solving of our problem.}, where $W$ and $H$ are the width and height of the input image, respectively, we formulate the problem of LDAP as solving for optimal perturbation: 
\begin{equation}
    \delta^* = \mathop{\arg\min}\limits_{\delta} L_{\text{attack}}(I+\delta)+\lambda_1L_{\text{area}}(\delta)+\lambda_2L_{\text{texture}}(\delta),
\label{eq:total_loss}
\end{equation}
where $\lambda_1$, $\lambda_2$ are weights; $\delta\in R^{W\times H}$ is the adversarial perturbation containing all adversarial patches; $L_{\text{attack}}$ is the attack loss that ensures  attack performance in different attack tasks; $L_{\text{area}}$ is the area loss that constrains the area of patches; and $L_{\text{texture}}$ is the texture loss that constrains the texture of patches. 
By solving this joint problem, we can get a proper $\delta$ that can deceive the object detector with small and texture-consistent patches. 

\subsubsection{Introduce texture and mask layers}

In order to define the shapes, positions and textures of the patches clearly, we factor $\delta$ into an element-wise multiplication of a texture layer $t \in R^{W\times H}$ and a mask layer $M \in{\{0,1\}}^{W\times H}$. 
The texture layer encodes the textures of patches, while the mask layer encodes the shapes and positions of patches. We divide the patches and the other regions with this binary mask layer $M$; the regions of patches in $M$ are set to 1 and the others are set to 0.
Then the perturbation~(the aggregation of all patches) can be cast as:
\begin{equation}
    \delta = t \odot M,
\end{equation}
where $\odot$ denotes the element-wise multiplication. One advantage of this factoring is that it separates out the shape and position terms and the texture term of the patches, making it easy to define $L_{\text{area}}$ directly on $M$.

Based on the definition of mask layer $M$ and texture layer $t$, the overall low-detectable attack problem is reformulated as:
\begin{equation}
\begin{aligned}
    &\min_{M,t} L_{\text{attack}}(I+t \odot M) + \lambda_1 L_{\text{area}}(M) + \lambda_2 L_{\text{texture}}(t).
\end{aligned}
\label{eq:total_loss_Mt}
\end{equation}
In this formulation, the area loss and texture loss are only related to $M$ and $t$, respectively, which makes it easier to separately optimize shapes and positions of patches, and textures of patches. This separation help us overcome the problem of coupling these two parts during optimization\footnote{The coupling means that the values of shapes and positions will influence the optimal values of the textures and vice versa. 
If we update both sides at the same time, the optimization can be unstable.
}.

\begin{figure}[t]
\centering
\includegraphics[width=1\linewidth]{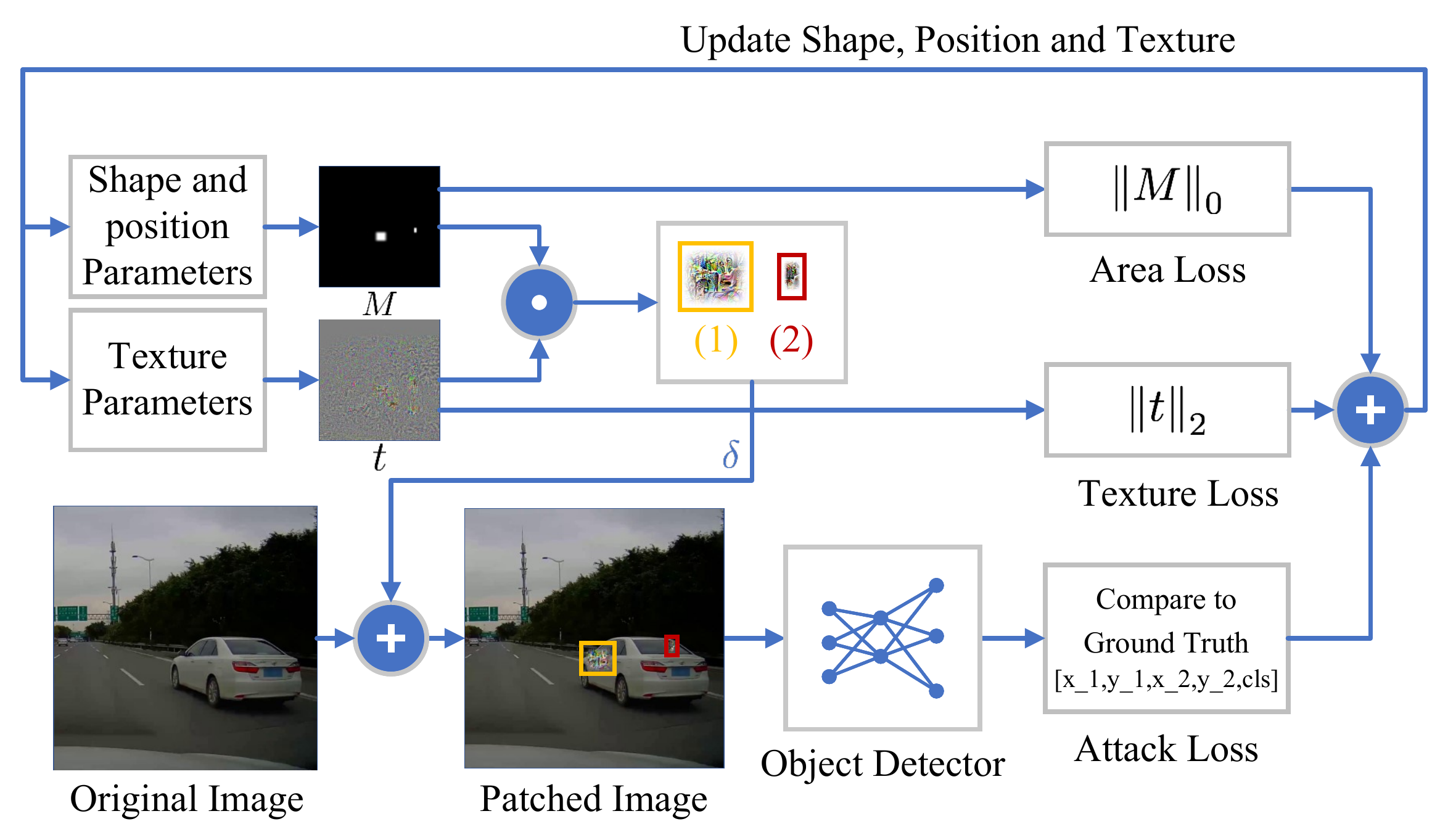}
\caption{ \small
The overview of our attack. The adversarial patches are calculated from the element-wise product of the mask layer~$M$ and the texture layer~$t$.
The area loss and the texture loss are directly calculated from $M$ and $t$, respectively. 
The attack loss is calculated by comparing the predictions of the object detector with the ground truth of the target object. 
We use the sum of the area loss, texture loss and attack loss to update the shape, position and texture of patches. 
}
\label{fig:method_overview}
\end{figure}

The overview of our attack method is shown in Fig.~\ref{fig:method_overview}. Given an input image, we first generate patches from shape, position and texture parameters, then add them onto the input image. The patched image is then passed into the object detector to calculate the attack loss. The area loss and the texture loss are computed directly from $M$ and $t$. Then we sum the loss functions and back-propagate to update the shape, position and texture parameters.
In the following three sections, we introduce these loss functions in detail.

\subsection{Attack Loss}
\label{sec:Attack_Loss}
First, we introduce the attack-loss function, which aims to corrupt the prediction of the detector. 
We introduce the soft-attack strategy we adopt in attack-loss functions and then the different attack tasks, followed by their corresponding attack-loss functions.

\subsubsection{Soft-attack strategy}
Previous attack methods adopt two common strategies when dealing with the redundant bounding boxes predicted on the target object. 
The first strategy is to attack only the bounding box with the maximum confidence score of the target category~\cite{liu2019dpatch,eykholt2018physical,thys2019fooling}. However, during attack optimization, the bounding box with the maximum confidence score could be changed from one bounding box to another continuously during optimization. This not only slows down the optimization, but also leads to inferior positions of patches and worse attack performance, resulting in larger patches, lower texture consistency and higher detectability.

The second strategy is to equally attack the bounding boxes that can be detected as the target object~\cite{huang2020universal}, which means that the predicted category scores of these bounding boxes---and the IoU between them and the target object---are higher than the thresholds of the detection model. This solution can alleviate the drawback of the first strategy, for these bounding boxes usually change less frequently than the bounding box with the maximum confidence score. However, most of these bounding boxes are not accurate, even though they are treated equally in this strategy. In our method, this leads to sub-optimal positions of patches and worse attack performance.

Therefore, we use a soft-attack strategy in our attack-loss function instead of these attack strategies.
The soft-attack strategy means that we assign different weights to the bounding boxes that can be detected as the target object. Here, we have an assumption that the bounding boxes closer to the target object are more worth attacking, for they are more likely to be the correct predictions. Based on this assumption, we set the IoU between the predicted bounding boxes and the bounding box of target object as the weights of each bounding box in the attack-loss function.
The soft-attack strategy concentrates our LDAP on the target object, and help us reduce the area of patches needed to deceive the detector.

\subsubsection{Different attack-loss functions}
\label{sec:Different_Attack_Loss}
Different from the classifier, an object detector incorporates two parts: object classification and localization. This makes the attack on a detector more complicated. Correspondingly, we set a classification-attack task and a localization-attack task that only focus on one function. For different tasks, we set different attack-loss functions.

Classification attack means that the adversary should make the detection model classify the target object as a wrong category. If we regard ``background'' as a special category, then an object's disappearance is also a special case of classification attack. 
In this paper, we study a danger for autonomous driving: a disappearing attack, which is an important attack task for not sighting a pedestrian ahead.
For a classification attack, we set the weighted sum of category scores as an attack loss. The loss function is:
\begin{align}
    L_\text{cls}(I+t \odot M) = \sum_{b_i\in\mathcal{B^*}} w_i C(I+t \odot M, b_i),
\end{align}
where $\mathcal{B^*}$ is the bounding boxes that could be detected as a target object; $b_i$ is the \emph{i}-th bounding box in $\mathcal{B^*}$; $b_t$ is the bounding box of the target object; $w_i = \text{IoU}(b_i, b_t)$ is the weight of $b_i$, which is assigned by our soft-attack strategy; and $C(\cdot)$ is the category score of the target category for $b_i$ predicted by the object detector on patched image $I+t \odot M$.

Localization attack means that the adversary should make the detection model predict bounding boxes with undesirable shape or position on the target object. 
In this paper, we investigate the attack on a bounding-box position, aiming to shift the bounding box horizontally. This attack task is important due to its potential threat to autonomous vehicles. For instance, if a bus ahead of an autonomous car is wrongly detected as being in the left lane, the car might drive forward and hit the bus.
In a localization attack, we set the weighted sum of regression offsets of bounding boxes as an attack loss. The loss function is:
\begin{align}
    L_\text{loc}(I+t \odot M) = \sum_{b_i\in\mathcal{B^*}} w_i R_x(I+t \odot M, b_i),
\end{align}
where $\mathcal{B^*}$ is the bounding boxes that could be detected as a target object; $b_i$ is the \emph{i}-th bounding box in $\mathcal{B^*}$; $b_t$ is the bounding box of the target object; $w_i = \text{IoU}(b_i, b_t)$ is the weight of $b_i$; and $R_x(\cdot)$ is the $x$-dimensional regression offset of $b_i$.

Therefore, the attack loss $L_{\text{attack}}$ is defined as: 
\begin{equation}
\begin{aligned}
& L_{\text{attack}} = 
    \begin{cases}
        L_\text{cls}, & \text{if a classification attack} \\
        L_\text{loc}, & \text{if a localization attack}.
    \end{cases}
\end{aligned}
\label{eq:attack_loss}
\end{equation}

\subsection{Area Loss}
\label{sec:Area_Loss}
It is easy to see that the area loss can be defined as
\begin{equation}
    L_{\text{area}} = {\|M\|}_0.
\label{eq:region_loss_M}
\end{equation}
However, directly solving for $M$ is a non-trivial high-dimensional discrete-search problem, since $M$ consists of $W \times H$ binary elements. ($W$ and $H$ are the width and height of the original image, respectively).
To circumvent this non-trivial problem, while referring to the aforementioned shape-modeling research~\cite{da2010fitting,panagiotakis2016parameter}, we use a set of geometric primitives to represent the mask layer.

\begin{figure}[t]
\begin{center}
\includegraphics[width=0.9\linewidth]{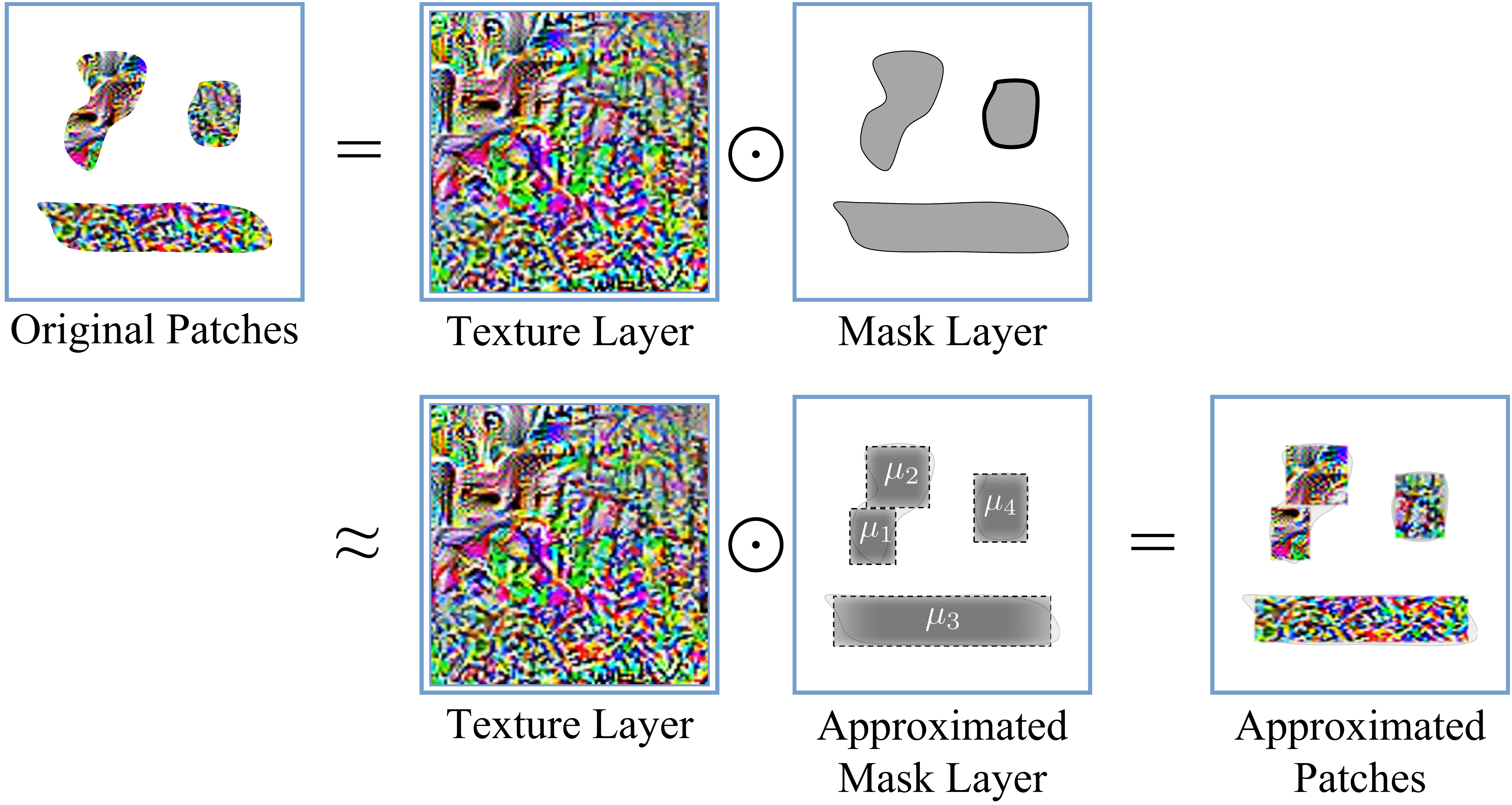}
\end{center}
\caption{
\small
Patch in irregular shapes and positions can be approximated by several rectangles. 
}
\label{fig:rand_approximate_big}
\end{figure}

Noticing that patches in irregular shapes and positions can be approximated by several rectangles, as shown in Fig.~\ref{fig:rand_approximate_big}, we make a reasonable assumption that the mask layer in our attack can be approximated by a set of geometric primitives. 
Therefore, we represent the mask layer as:
\begin{equation}
M = \bigcup^N_j \mu_j,
\label{eq:M_and_mu}
\end{equation}
where $\mu_j \in {\{0,1\}}^{W \times H}$ is the $j$-th geometric primitive, and $N$ is the number of primitives.
Each primitive can be regarded as a degraded mask layer, which encodes only one part of the whole mask layer with a simple shape.
In this way, we convert the optimization of $M$ to an optimization of parameters of primitives, dramatically reducing the number of parameters to optimize.

\begin{figure}[t]
\centering
\includegraphics[width=0.5\linewidth]{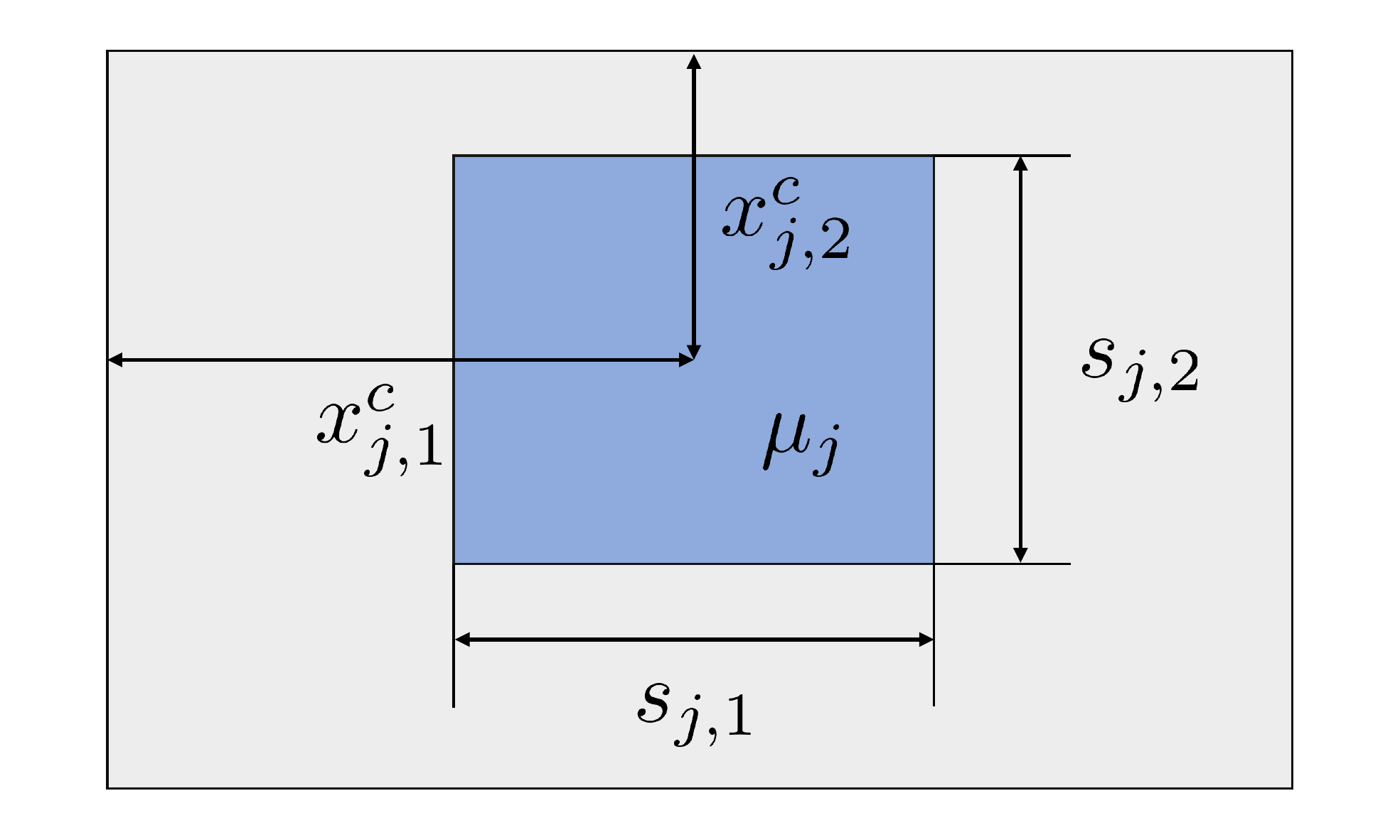} 
\caption{
\small
Parameters of $j$-th rectangle base $\mu_j$. The blue region is filled with 1, and the gray region is padded with 0.}
\label{fig:block_param}
\end{figure}

Without loss of generality, we first choose rectangles as geometric primitives, 
a reasonable choice because both the receptive fields and the predicted bounding box of detection models are rectangles. 
Almost all previous patch-wise attacks use rectangle patches, such as those of DPatch~\cite{liu2019dpatch}, AdvPatch~\cite{thys2019fooling}, $RP_2$-Sticker~\cite{eykholt2018physical} and UPC~\cite{huang2020universal}. The rectangle primitive is written as:
\begin{equation}
\begin{aligned}
    \mu_j(\bm{x};\bm{x^c_j},\bm{s_j})= 1, \text{if $|x_k - x^c_{j,k}| \leq \frac{s_{j,k}}{2}$, $k=1,2$},
\end{aligned}
\label{eq:uniform_value_rectangle}
\end{equation}
where $\bm{x}=(x_1, x_2)$ denotes the coordinates of pixels in the image; $\bm{x^c_j}=(x^c_{j,1}; x^c_{j,2})$ denotes the position parameters~(coordinates of the center point) of the $j$-th rectangle; and $\bm{s_j}=(s_{j,1}; s_{j,2})$ denotes its shape parameters~(width and height). 
The demonstration of a rectangle primitive $\mu_j$ with its parameters is shown in Fig.~\ref{fig:block_param}.

\subsubsection{Solving Non-differential Problem}

However, the rectangle primitive defined in Eq.~\eqref{eq:uniform_value_rectangle} induces a steep change from 1 to 0 on the boundary of the rectangle, resulting in non-differentiability of the loss function with respect to the position and shape parameters of the primitives $\bm{x^c_j}$ and $\bm{s_j}$. 
To address this non-differentiability issue, we relax the boundary of a rectangle primitive using a cosine function.
The boundary-relaxed primitive is: 
\begin{equation}
\begin{aligned}
& \mu_j(\bm{x}; \bm{x^c_j}, \bm{s_j}) = 
\phi( \frac{2\pi}{s_{j,1}}(x_1-x^c_{j,1}))
\phi( \frac{2\pi}{s_{j,2}}(x_2-x^c_{j,2})),
\\ & \text{where } \phi(z) = 
    \begin{cases}
        \frac{1}{2}(\cos(z+\frac{\pi}{2}) + 1), & \text{if $-\frac{3\pi}{2}<z<-\frac{\pi}{2}$} \\
        1, & \text{if $-\frac{\pi}{2}<z<\frac{\pi}{2}$} \\
        \frac{1}{2}(\cos(z-\frac{\pi}{2}) + 1), & \text{if $\frac{\pi}{2}<z<\frac{3\pi}{2}$} \\
    	0, & \text{else}
    \end{cases}
    .
\end{aligned}
\label{eq:cos_bases}
\end{equation}
We find that the cosine-boundary primitive obtains a smooth boundary, continuous at all points, making the loss function differentiable with respect to its position and shape parameters.

After the relaxation of the boundary, the value range of a primitive changes from discrete to continuous. Therefore, we rewrite the union of primitives in Eq.~\eqref{eq:M_and_mu} as a weighted sum of primitives such that:
\begin{equation}
M = \sum^N_j \alpha_j \mu_i(\bm{x^c_j}, \bm{s_j}),
\label{eq:M_and_mu_new}
\end{equation}
where $\alpha_j$ is the weight we set for the $j$-th primitive, and $\alpha_j>0$. This rewriting implies an assumption that the primitives do not overlap each other in the mask layer. This assumption is reasonable because \textbf{(i)}~the overlap is a waste of the area of primitives, and \textbf{(ii)}~the overlapped primitives can be replaced equivalently with non-overlapped primitives.

Based on Eq.~\eqref{eq:M_and_mu_new}, we can expediently convert the $l_0$ norm of $M$ in Eq.~\eqref{eq:region_loss_M}, which is difficult to optimize, to the sum of the area of each primitive. The area loss is:
\begin{equation}
    L_{\text{area}} = \sum^N_j s_{j,1} s_{j,2}, 
\label{eq:region_loss}
\end{equation}
where $s_{j,1}$ and $s_{j,2}$ are the width and height of a $j$-th rectangle primitive.

\subsection{Texture Loss}
\label{sec:Texture_Loss}
Texture loss aims to maintain consistency between the texture of patches and that of the original image.
Therefore, we directly use the $l_2$ norm of adversarial texture $t$ as our texture loss. The texture loss is:
\begin{equation}
    L_{\text{texture}}(\delta) = {\|t \|}_2.
\label{eq:texture_loss}
\end{equation}

Finally, we substitute Eq.~\eqref{eq:attack_loss},
Eq.~\eqref{eq:cos_bases},
Eq.~\eqref{eq:M_and_mu_new},
Eq.~\eqref{eq:region_loss} and Eq.~\eqref{eq:texture_loss} to Eq.~\eqref{eq:total_loss_Mt} to calculate the adversarial perturbation $\delta$, which is the aggregation of all patches.

\subsection{Optimization}
\label{sec:Optimization}

As we mentioned before, the optimization of $M(\bm{x^c},\bm{s}, \bm{\alpha})$ and $t$ in the problem defined in Eq.~\eqref{eq:total_loss_Mt} are coupled, which can destabilize our optimization and cause sub-optimal attack performance. Here $\bm{x^c} = \{\bm{x^c_j}\}$, $\bm{s}=\{\bm{s_j}\}$, $\bm{\alpha}=\{\bm{\alpha_j}\}$, and $j=1,\ldots, N$.
Therefore, we separate the problem into two separated aspects, namely a \emph{Mask Layer Search} and a \emph{Texture Layer Search}, and alternately optimize $M$ or $t$ step-by-step, to indirectly solve the problem.

\subsubsection{Mask Layer Search}
During this search step, we fix the texture $t$ and optimize the mask layer $M$ controlled by $\bm{x^c}$, $\bm{s}$ and $\bm{\alpha}$. 
Due to the separation of the mask layer and texture layer, we can directly omit the texture loss in this step.
The problem in this step is:
\begin{equation}
    \min_{\bm{x^c},\bm{s}, \bm{\alpha}} L_{\text{attack}}(I + t \odot M(\bm{x^c},\bm{s},\bm{\alpha})) + \lambda_1 L_{\text{area}}(\bm{s}).
\label{eq:attack6}
\end{equation}

Note that during optimization, $\bm{\alpha}$ is coupled with $\bm{x^c}$ and $\bm{s}$, such that we can separate the updates of $\bm{x^c}$, $\bm{s}$ and $\bm{\alpha}$. 
We denote the objective function defined in Eq.~\eqref{eq:attack6} as $J$.
First we fix $\bm{\alpha}$ and calculate the gradient of $J$ with respect to $\bm{x^c}$ and $\bm{s}$ to update them,
\begin{equation}
    \bm{x^c} \gets \bm{x^c} - \eta_x \frac{\partial J}{\partial \bm{x^c}},
    \bm{s} \gets \bm{s} - \eta_s \frac{\partial J}{\partial \bm{s}},
\label{eq:update_xS}
\end{equation}
where $\eta_x$, $\eta_s$ are the learning rates of $\bm{x^c}$ and $\bm{s}$, respectively.
Then we fix $\bm{x^c}$ and $\bm{s}$ and update $\bm{\alpha}$,
\begin{equation}
    \bm{\alpha} \gets \bm{\alpha} - \eta_\alpha \frac{\partial J}{\partial \bm{\alpha}},
\label{eq:update_alpha}
\end{equation}
where $\eta_\alpha$ is the learning rate of $\bm{\alpha}$.
After these two updates, we finish one region-search step, in which we update the mask layer $M$ based on the fixed texture layer.

\subsubsection{Texture Layer Search}
After one step of a region search, we fix the mask $M$ and update the texture $t$ for one step. Ignoring the terms that are not related to $t$ in Eq.~\eqref{eq:total_loss_Mt}, we write the texture-layer search problem as:
\begin{equation}
    \min_{t} L_{\text{attack}}(I + t \odot M(\bm{x^c},\bm{s},\bm{\alpha})) + \lambda_2 L_{\text{texture}}(t).
\label{eq:texture_search_0}
\end{equation}
We also calculate the gradient to update the texture $t$ for one step, such that: 
\begin{equation}
    t \gets t - \eta_t \frac{\partial J^\prime}{\partial t},
\label{eq:update_t}
\end{equation}
where $\eta_t$ is the learning rate of $t$, and $J^\prime$ is the objective function of Eq.~\eqref{eq:texture_search_0}. In this step, we update the texture layer based on the fixed-mask layer.

\subsubsection{Increasing $\lambda_1$ and $\lambda_2$}

For different images, the weights of area loss and texture loss should not be always the same, since objects in different images are usually not the same size (smaller objects usually need smaller patches to attack, so the weight of area loss should be bigger). To overcome this problem, we set the value of $\lambda_1$ and $\lambda_2$ to gradually increase after the texture-search step. This setting can ensure that, for different images, LDAP can find small enough and texture-consistent patches, 
\begin{equation}
\lambda_1 \gets \lambda_1 + \Delta \lambda, \lambda_2 \gets \lambda_2 + \Delta \lambda,
\label{eq:update_lambda}
\end{equation}
where $\Delta \lambda$ is the increasing rate of $\lambda_1$ and $\lambda_2$.
After this step, we turn to the next step of region search. 
The overall procedure of our method is illustrated in Algorithm~\ref{alg:ldap}.

\begin{figure}[!t]
  \renewcommand{\algorithmicrequire}{\textbf{Input:}}
  \renewcommand{\algorithmicensure}{\textbf{Output:}}
  \begin{algorithm}[H]
    \small
    \caption{Algorithm of LDAP for One Image} 
    \begin{algorithmic}[1] 
    
    \REQUIRE
    Image $I$;
    Number of rectangle primitives $N$;
    bounding box of target object $b_t$;
    learning rates $\eta_x$, $\eta_s$, $\eta_\alpha$, $\eta_t$;
    initial coefficients $\lambda_1, \lambda_2$ and their increasing rate $\Delta \lambda$;
    maximum search step $n_{\mathrm{max}}$.
    
    \ENSURE 
    Adversarial perturbation $\delta$:
    \STATE
    initialize texture layer $t$, weights of bases $\bm{\alpha}$, position and shape parameters of primitives $\bm{x^c}, \bm{s}$
    \STATE
    step $\gets 0$
    \WHILE{step $< n_{\mathrm{max}}$}     
        \STATE
        search mask layer; update $\bm{x^c}$ and $\bm{s}$ according to Eq.~\eqref{eq:update_xS}
        \STATE
        search mask layer; update $\bm{\alpha}$ according to Eq.~\eqref{eq:update_alpha}
        \STATE
        search texture layer; update $t$ according to Eq.~\eqref{eq:update_t}
        \STATE
        increase $\lambda_1,\lambda_2$ according to Eq.~\eqref{eq:update_lambda}
        \STATE
        step $\gets$ step $+ 1$
        
    \ENDWHILE
    \STATE
    compute perturbation $\delta \gets M(\bm{x^c},\bm{s},\bm{\alpha}) \odot t$
    \end{algorithmic}
    \label{alg:ldap}
    \end{algorithm}
\end{figure}

\section{Experiments}
\label{sec:Experiments}
In this section, we experimentally evaluate our approach's detection-attacking effectiveness, and the resulting threat to autonomous vehicles. 
In Section~\ref{sec:Experiment_Settings}, we introduce the experiment settings. 
In Section~\ref{sec:Experiments_on_COCO}, we verify the effectiveness of our method compared to other attacks on COCO, which is a common detection dataset.
In Section~\ref{sec:Experiments_on_dd}, we demonstrate two potential attack threats in autonomous driving on $\text{D}^2$-City, which is a driving-record video dataset.
Then we do some ablation studies in Section~\ref{sec:Ablation_Study}.

\subsection{Experiment Settings}
\label{sec:Experiment_Settings}

Without loss of generality, we choose ``Person'' in COCO and ``Car'' in $\text{D}^2$-City as the target categories,  both important categories in the field of autonomous driving, and widely studied in previous research~\cite{thys2019fooling,xu2020adversarial,huang2020universal}.

\subsubsection{Datasets}
\begin{figure}[t]
\centering
\subfloat[]{
\begin{minipage}[]{.185\linewidth} 
\includegraphics[width=1\linewidth]{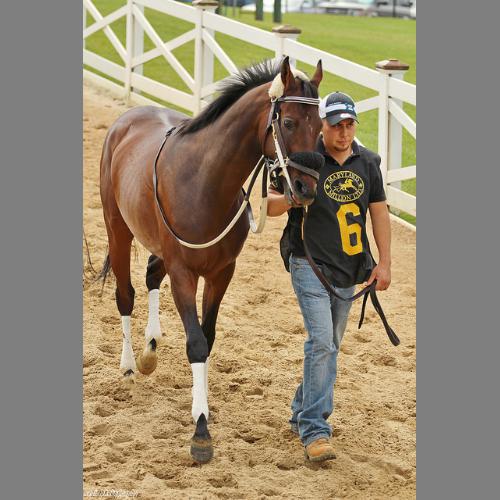}
\\\vspace{-9pt}
\includegraphics[width=1\linewidth]{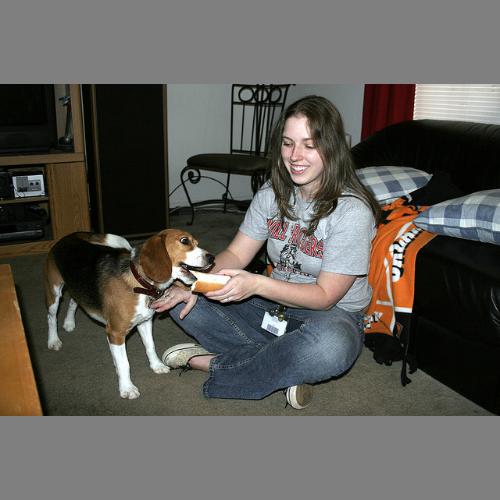}
\end{minipage}
}
\subfloat[]{
\begin{minipage}[]{.185\linewidth} 
\includegraphics[width=1\linewidth]{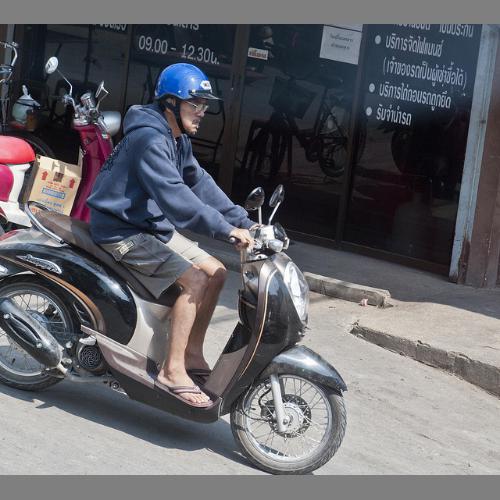}
\\\vspace{-9pt}
\includegraphics[width=1\linewidth]{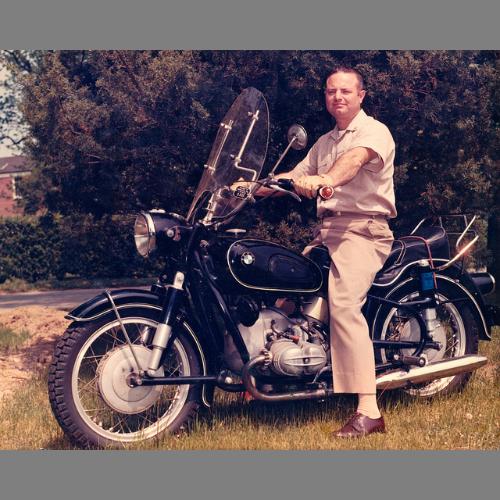}
\end{minipage}
}
\subfloat[]{
\begin{minipage}[]{.185\linewidth} 
\includegraphics[width=1\linewidth]{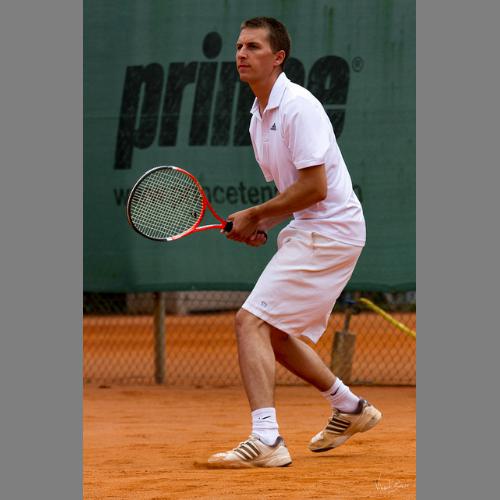}
\\\vspace{-9pt}
\includegraphics[width=1\linewidth]{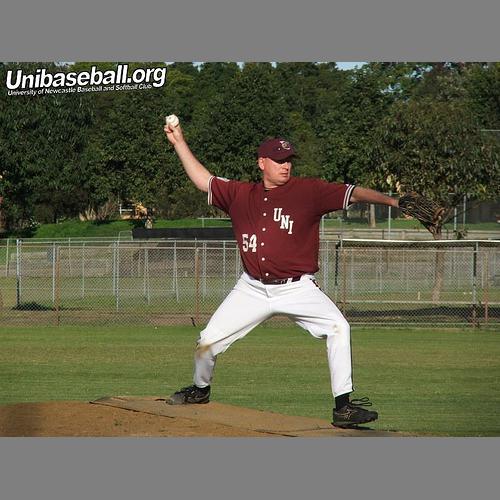}
\end{minipage}
}
\subfloat[]{
\begin{minipage}[]{.185\linewidth} 
\includegraphics[width=1\linewidth]{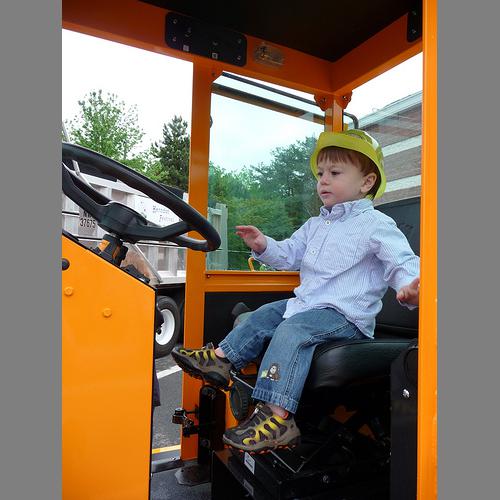}
\\\vspace{-9pt}
\includegraphics[width=1\linewidth]{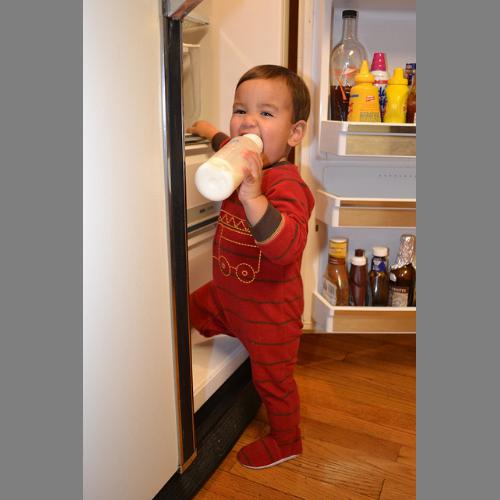}
\end{minipage}
}
\subfloat[]{
\begin{minipage}[]{.185\linewidth} 
\includegraphics[width=1\linewidth]{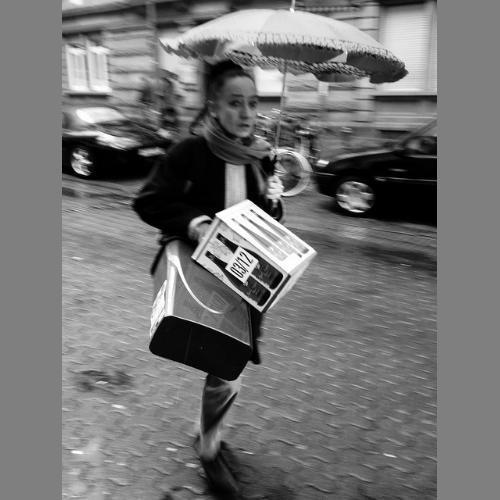}
\\\vspace{-9pt}
\includegraphics[width=1\linewidth]{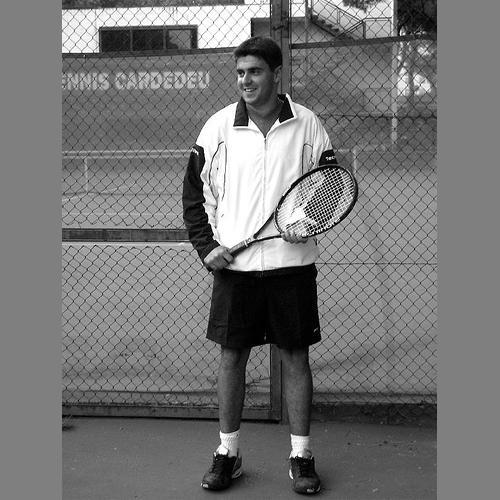}
\end{minipage}
}
\caption{
Images selected from the COCO dataset:
(a)~people with animals;
(b)~people with motorbikes;
(c)~people playing sports;
(d)~children; and
(e)~people in black-and-white photos.
}
\label{fig:coco_dataset}
\end{figure}

\begin{figure}[t]
\centering
\subfloat[]{
\begin{minipage}[]{.185\linewidth} 
\includegraphics[width=\linewidth]{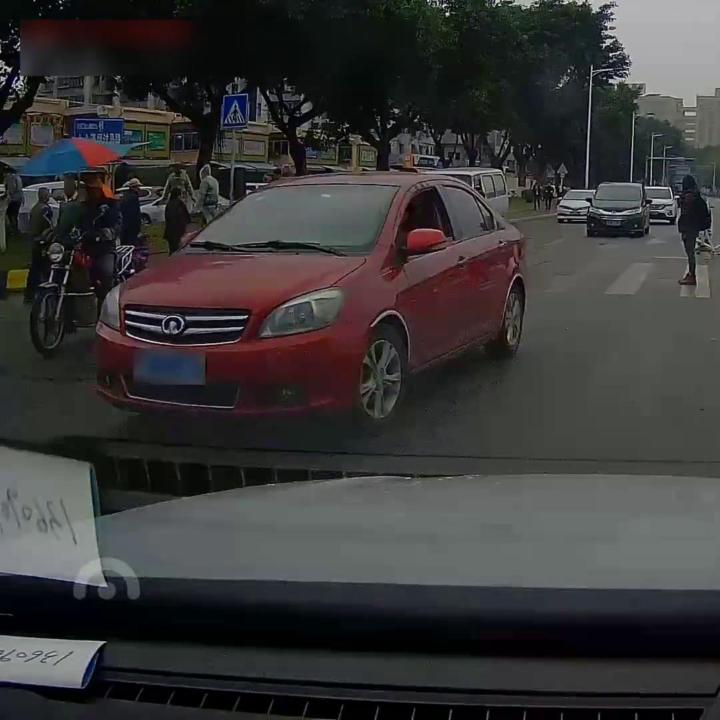}
\\\vspace{-9pt}
\includegraphics[width=\linewidth]{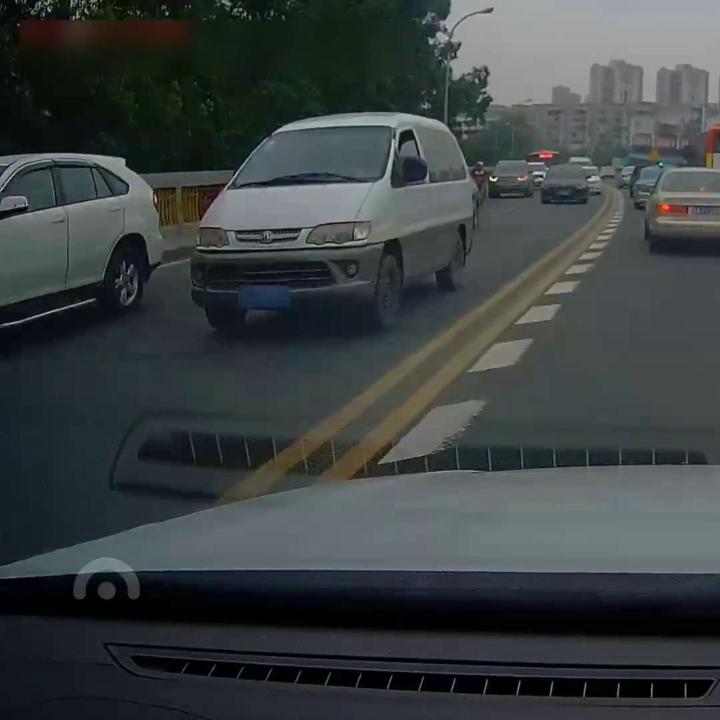}
\end{minipage}
}
\subfloat[]{
\begin{minipage}[]{.185\linewidth} 
\includegraphics[width=\linewidth]{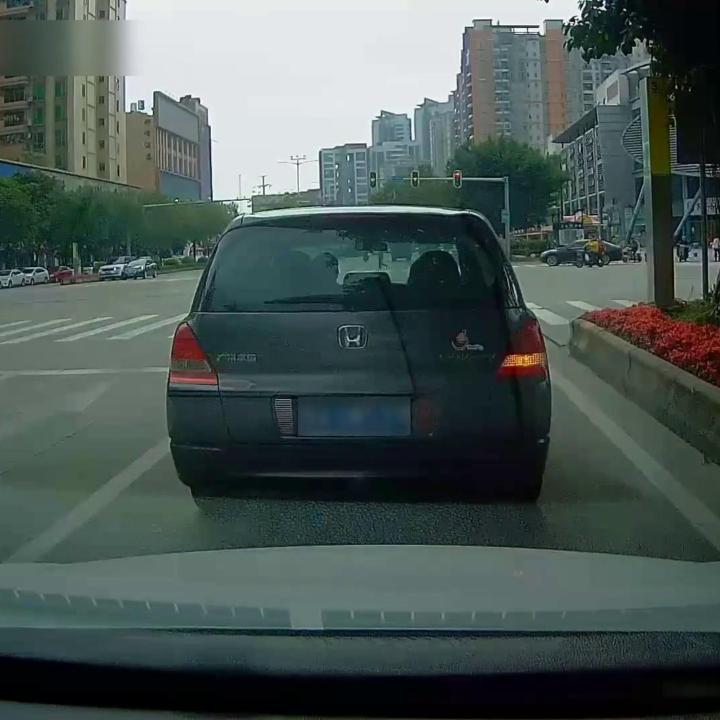}
\\\vspace{-9pt}
\includegraphics[width=\linewidth]{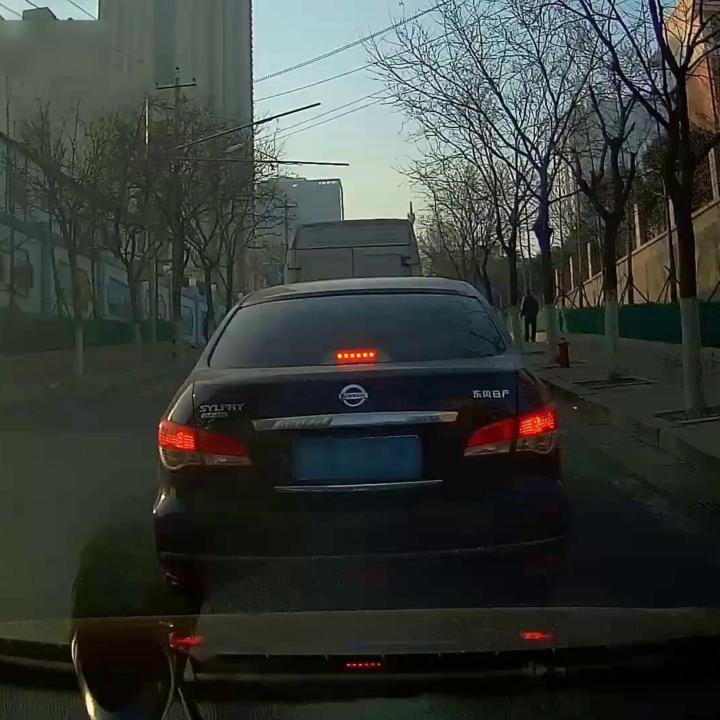}
\end{minipage}
}
\subfloat[]{
\begin{minipage}[]{.185\linewidth} 
\includegraphics[width=\linewidth]{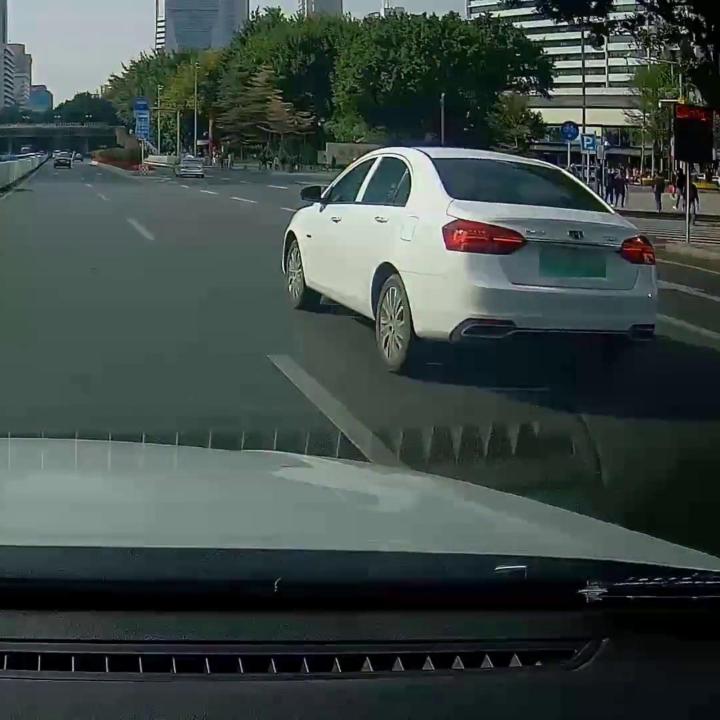}
\\\vspace{-9pt}
\includegraphics[width=\linewidth]{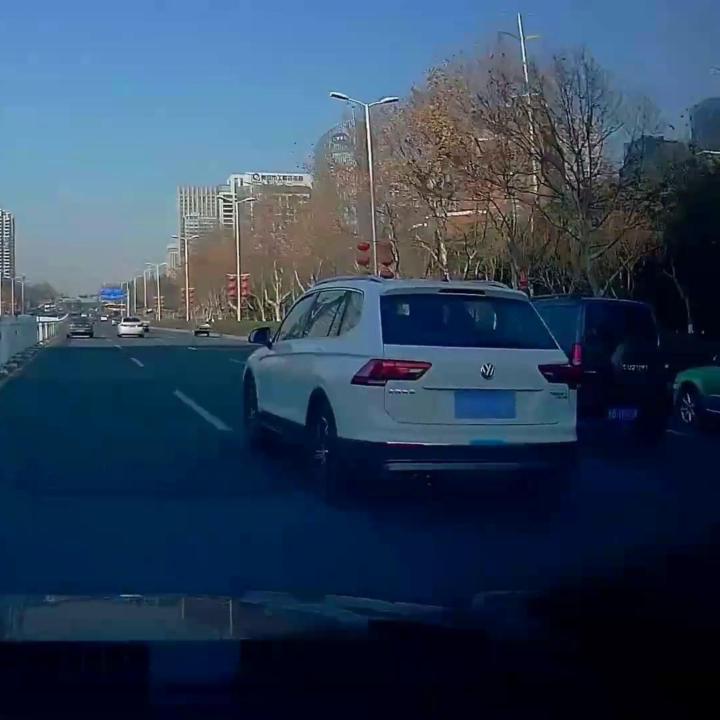}
\end{minipage}
}
\subfloat[]{
\begin{minipage}[]{.185\linewidth} 
\includegraphics[width=\linewidth]{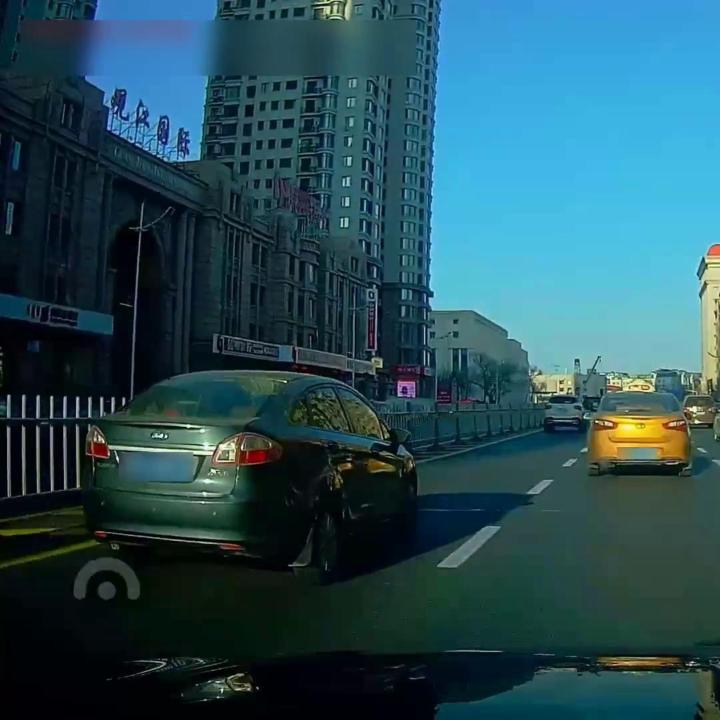}
\\\vspace{-9pt}
\includegraphics[width=\linewidth]{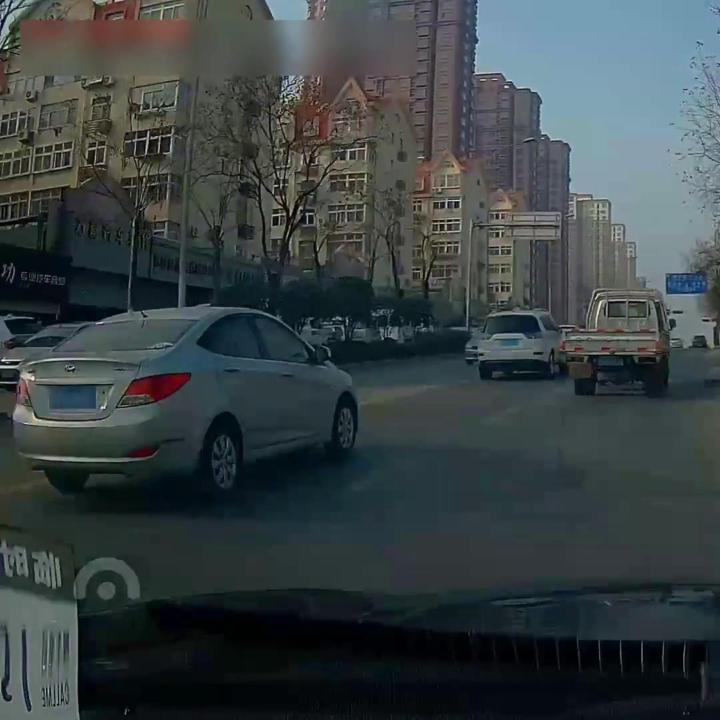}
\end{minipage}
}
\subfloat[]{
\begin{minipage}[]{.185\linewidth} 
\includegraphics[width=\linewidth]{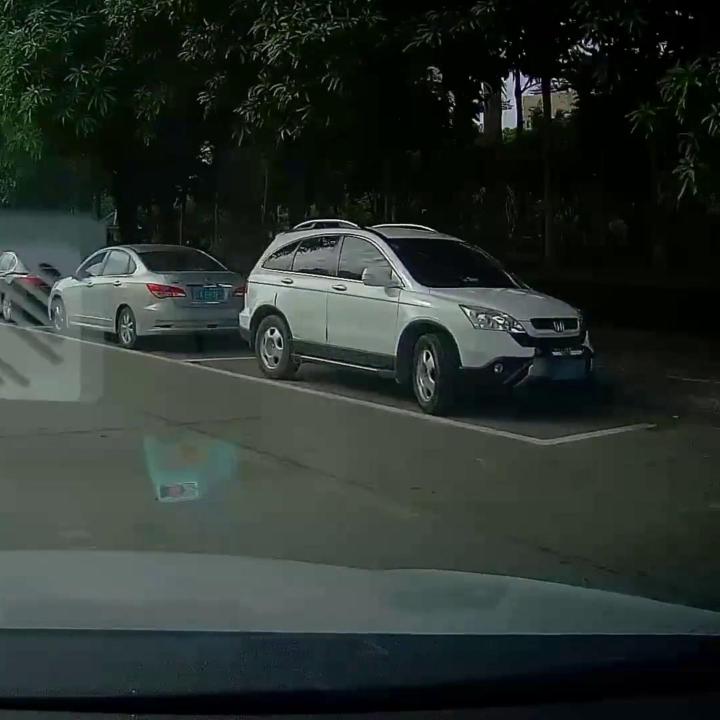}
\\\vspace{-9pt}
\includegraphics[width=\linewidth]{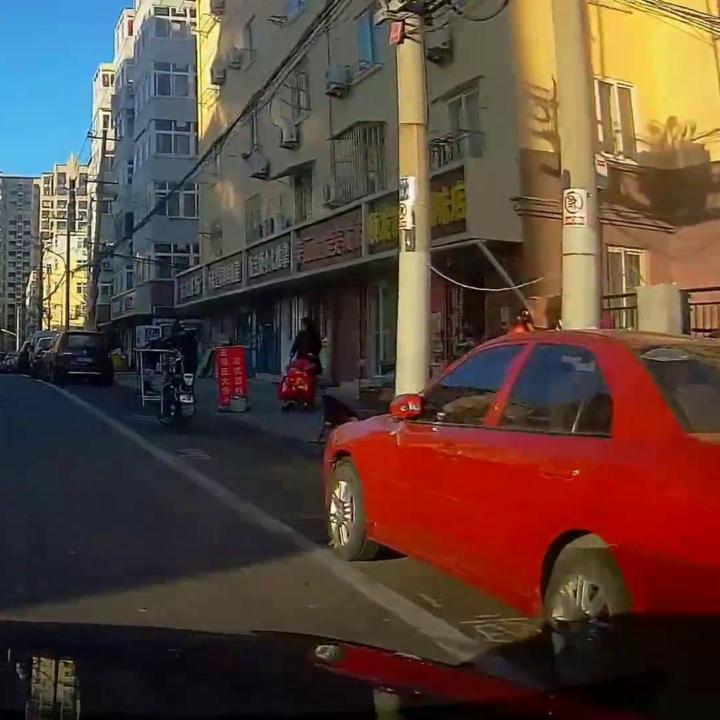}
\end{minipage}
}
\caption{
Images selected from the $\text{D}^2$-City dataset:
(a)~opposite cars;
(b)~cars directly in front;
(c)~cars on the right-front side;
(d)~cars on the left-front side; and
(e)~parked cars.
}
\label{fig:d2-city_dataset}
\end{figure}

We use two datasets for different purposes. To validate the effectiveness and superiority of our method in a common detection scenario, we select images from the widely used detection dataset COCO. To demonstrate the threats of our method in an autonomous driving scenario, we select images from the driving-record video dataset $\text{D}^2$-City. 
\begin{itemize}

    \item \textbf{COCO} is a large dataset for both object detection and segmentation, containing more than 330,000 images~\cite{lin2014microsoft}. 
    We select 1,454 person images from the COCO 2017 training set. 
    Some of the images we selected from the COCO dataset are shown in Fig~\ref{fig:coco_dataset}.

    \item \textbf{$\bm{\text{D}^2}$-City} is a driving-record video dataset collected in five cities, and containing 355 video segments. We sample the videos at the interval of one per second, then select 1,096 frames that contain cars from the train set of $\text{D}^2$-City. 
    Some of the images that we selected are shown in Fig~\ref{fig:d2-city_dataset}.

\end{itemize}

All of the target objects in the selected images satisfy the following criteria: \textbf{(i)}~ large enough; \textbf{(ii)}~ correctly detected by all detectors we test on; and \textbf{(iii)}~not occluded by other objects. All of these criteria ensure the difficulty of attacking detectors. 

\subsubsection{Detection Models}
We attack four representative object detection models in our experiments: two-stage models including Faster R-CNN (FRCN)~\cite{ren2015faster} and Mask R-CNN (MRCN)~\cite{he2017mask}, and one-stage models including SSD~\cite{liu2016ssd} and YOLOv3~\cite{redmon2018yolov3}. Specifically, Faster R-CNN and Mask R-CNN use Resnet-50~\cite{he2016deep} and FPN~\cite{lin2017feature} as backbones, while SSD and YOLOv3 use VGG-16~\cite{simonyan2014very} and Darknet-53 as backbones, respectively. All of these detectors are implemented by mmdetection~\cite{chen2019mmdetection}\footnote{https://github.com/open-mmlab/mmdetection}.

\subsubsection{Parameters Setting}
We set the initial value of $\alpha_1,\alpha_2$ to 1e-2 and the increasing rate $\Delta \lambda$ to 1e-4; the maximum search step $n_{\mathrm{max}}$ to 2,000; and the number of rectangle primitives $N$ to 10.

\subsubsection{Comparison Methods}
We compare our method with four former attack methods as well as with two attack methods designed by us. All comparable attack methods adopt fixed positions of patches. Furthermore, we reproduce the codes of all former attacks to fit the PyTorch and mmdetection frameworks.

\begin{itemize}
    \item \textbf{DPatch}~\cite{liu2019dpatch} aims to reduce the maximum category score of the target class in all bounding boxes with a rectangle patch in the upper-left corner of the input image. 

    \item \textbf{AdvPatch}~\cite{thys2019fooling} aims to reduce the maximum object-ness score in all predicted bounding boxes, using a rectangle patch in the center of the target object. However, reducing the object-ness score of the bounding box has nearly no influence on Faster R-CNN, Mask R-CNN and SSD; therefore, we change AdvPatch to reduce category scores in these models. 
    
    \item \textbf{$\bm{RP_2}$-Sticker}~\cite{eykholt2018physical} aims to reduce the maximum category score of the target class in all bounding boxes, using two rectangle patches in the centers of the upper- and lower-half of the target object.
    
    \item \textbf{UPC}~\cite{huang2020universal} aims to reduce the sum of the category scores of the target class of the bounding boxes that can be detected as the target object. We reproduce the code of the 7-Patterns version of UPC~(\textbf{UPC-7P}), which puts adversarial patterns on seven certain parts of the human body~(upper arms, thighs, calves and abdomen) to deceive the detector.
    
    \item \textbf{2-Rectangles-Horizontal}~(2-Rects-H) attack is the first comparison method designed by us. Unlike the $\bm{RP_2}$-Sticker, our 2-Rects-H puts two rectangle patches in the center of the left- and right-half of the target object, and also shares the attack loss of our LDAP.

    \item \textbf{4-Rectangles}~(4-Rects) attack is the second comparison method that we design, and puts four rectangle patches in the center of the upper-left, upper-right, lower-left and lower-right quarters of the target object, while also sharing the attack loss of our LDAP.

\end{itemize}

For all comparison methods, we start with small patches, and gradually increase the patches' size until the attack succeeds. With this setting, we can calculate the minimum region needed to deceive the detector of these methods.
Considering that the experiments are all conducted in digital settings, we remove the physical modules in AdvPatch, $RP_2$-Sticker and UPC-7P, such as Expectation of Transformation~\cite{athalye2018synthesizing}, image projection and printed color shift modeling.
Since the original DPatch, AdvPatch, $RP_2$-Sticker and UPC-7P are not able to perform a localization attack, we replace their loss functions with our attack-loss function in localization-attack task mode, but maintain their manipulated region settings.

\subsubsection{Adversarial Patch Detector}
\label{sec:Malicious_detection_module}
To the best of our knowledge, there is no proper adversarial patch-detection method for an object-detection attack. 
However, we find that existing patch-wise adversarial attacks are not only visually perceptible to humans, but also easy to detect by a simple two-category classification network.
With a small number of labeled pairs of adversarial examples as well as corresponding clean images, we can easily train a classifier to determine whether the input image is adversarial or not, with 90\% accuracy.

In our experiments for our adversarial patch detector, we use a ResNet-18 network~\cite{he2016deep}, followed by one fully connected layer---all very simple yet still effective for adversarial patch detection. 
With COCO, we use 500 adversarial examples and their corresponding 500 clean images for training, 200$\times$2 for validation and 500$\times$2 for testing. 
In $\text{D}^2$-City, we use 400$\times$2 images for training, 200$\times$2 for validation and 400$\times$2 for testing. 
To overcome the overfitting, we use the network with the best validation performance to test. 
Therefore, the detectability of different attack methods is measured by the accuracy of the adversarial patch detectors.

\subsubsection{Evaluation Metric}
For all detection models, we set the confidence threshold as 0.3, the same as the default threshold in mmdetection. 

\textbf{For attack performance}, we use Attack Success Rate~(ASR) under a specific attack area threshold to evaluate the attack performance under the same patch area constraint. For a certain image, the attack succeeds if \textbf{(i)}~the attack deceives the detector, and \textbf{(ii)}~the sum of the patches' area is lower than this threshold. The threshold will be specifically declared in different tasks for different detectors. The ASR is:
\begin{equation}
    \text{ASR} = \frac{N_A}{N_I},
\end{equation}
where $N_I$ denotes the number of images in the dataset, and $N_A$ denotes the number of successfully attacked images under the patch area constraint.

\textbf{For reducing patch area}, we define the Mean Area Ratio~(MAR) at ASR $\geq$ 95\% to evaluate the ability to reduce the patch area. 
MAR is calculated by the average ratio of manipulated area and object area over all successfully attacked objects in dataset, which is:
\begin{equation}
    \text{MAR} = \frac{1}{N_A}\sum_n \frac{R_{A,n}}{R_{O,n}},
\end{equation}
where $R_{A,n}$, $R_{O,n}$ are the summed areas of the patches as well as the target object region area on the $n$-th adversarial example, respectively.

\textbf{For texture consistency}, we use Learned Perceptual Image Patch Similarity~(LPIPS)~\cite{zhang2018unreasonable} as our texture consistency metric to evaluate the consistency between adversarial examples and benign images. LPIPS is a widely used metric to evaluate perceptual similarity between images. The LPIPS is defined as:
\begin{equation}
    \text{LPIPS} = \frac{1}{N_A} \sum_n \sum_l \frac{1}{H_lW_l}\sum_{h,w}{\left\|f^l_{hw}-f^l_{0hw}\right\|}_2^2,
\end{equation}
where $f^l_{hw}$ and $f^l_{0hw}$ are respectively the features of the $n$-th adversarial example and the corresponding benign image on the $l$-th layer of the VGG-network. The lower $\text{LPIPS}$ indicates higher texture consistency.

\textbf{For detectability}, we define Adversarial Detection Accuracy~(ADA) to evaluate the detectability of each attack.
\begin{equation}
\begin{aligned}
    &\text{ADA} = \frac{TP+TN}{N_A \times 2}, \\
\end{aligned}
\end{equation}
where $TP$ is the number of True Positive images, and $TN$ is the number of True Negative images. We multiply $N_A$ by 2 because we test the detection accuracy on both adversarial examples and benign images.

\subsection{Experiments on COCO}
\label{sec:Experiments_on_COCO}
To validate the effectiveness and superiority on attack performance, small patches, consistent texture and low detectability of LDAP, we do experiments on the commonly used detection-dataset COCO. For different attack tasks, we introduce the experiments separately.

\subsubsection{Classification Attack}
Classification attack means the attack should mislead the model to misclassify the target object.
As we discussed in Section~\ref{sec:Different_Attack_Loss}, a disappearing attack is an important classification-attack task to study, so we study it in this section.
The attack success criterion is that there is no bounding box predicted in the correct category that overlaps with the bounding box of the target object. This means that simply changing the size of a bounding box (not making it disappear) is not considered to be a successful disappearing attack in our problem setting.

\begin{figure}[t!]
\centering 
\begin{minipage}[]{.19\linewidth} \scriptsize \centering
\subfloat[]{
\includegraphics[width=\linewidth]{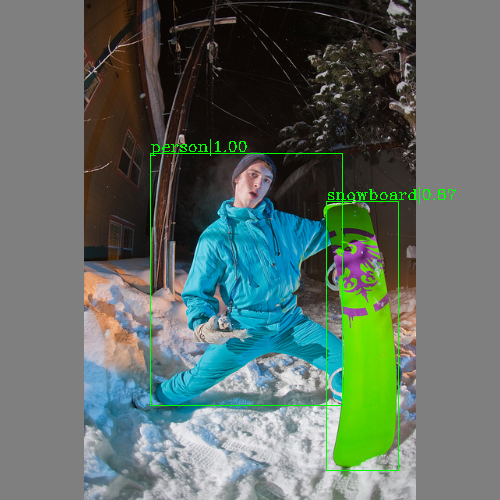}
}
\end{minipage}
\begin{minipage}[]{.6\linewidth} \scriptsize \centering
\subfloat[]{
\includegraphics[width=.32\linewidth]{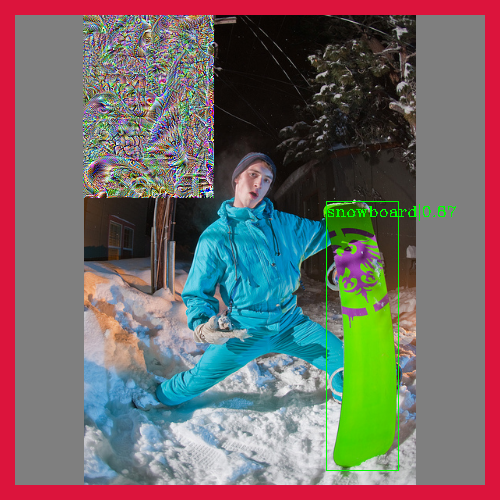}
}
\subfloat[]{
\includegraphics[width=.32\linewidth]{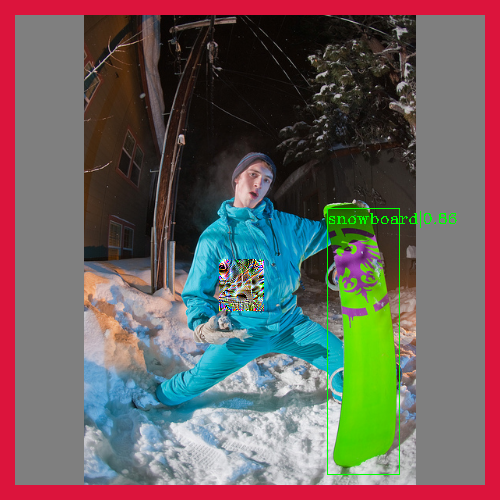}
}
\subfloat[]{
\includegraphics[width=.32\linewidth]{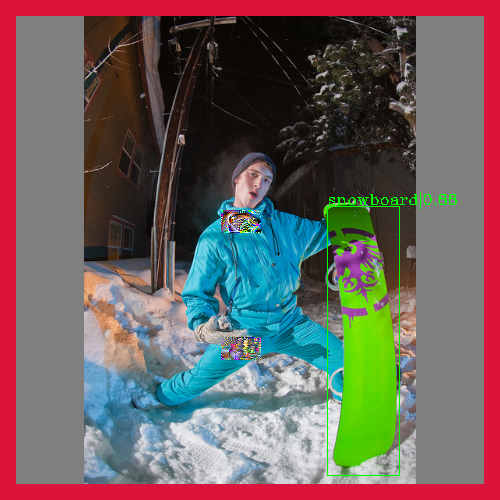}
}
\\
\subfloat[]{
\includegraphics[width=.32\linewidth]{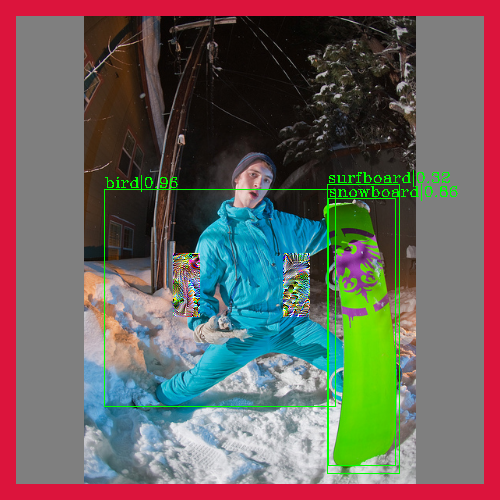}
}
\subfloat[]{
\includegraphics[width=.32\linewidth]{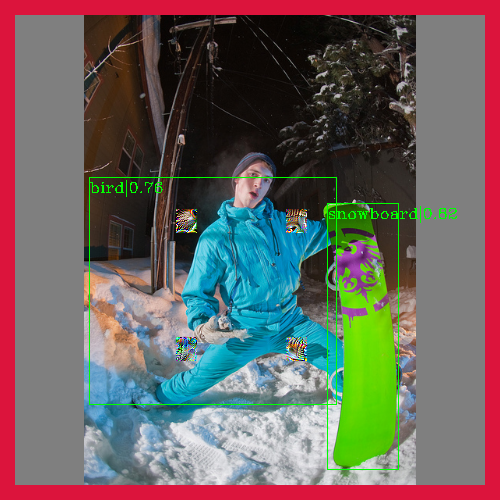}
}
\subfloat[]{
\includegraphics[width=.32\linewidth]{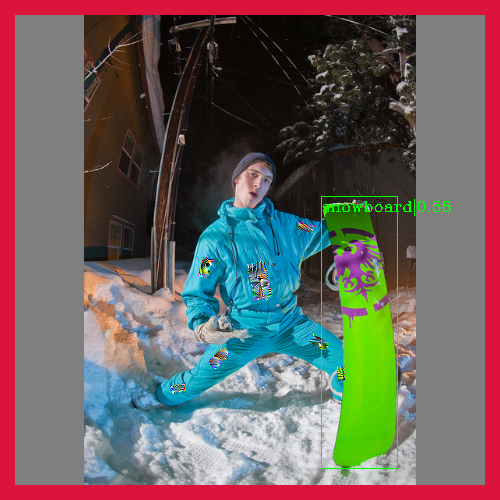}
}
\end{minipage}
\begin{minipage}[]{.19\linewidth} \scriptsize \centering
\subfloat[]{
\includegraphics[width=\linewidth]{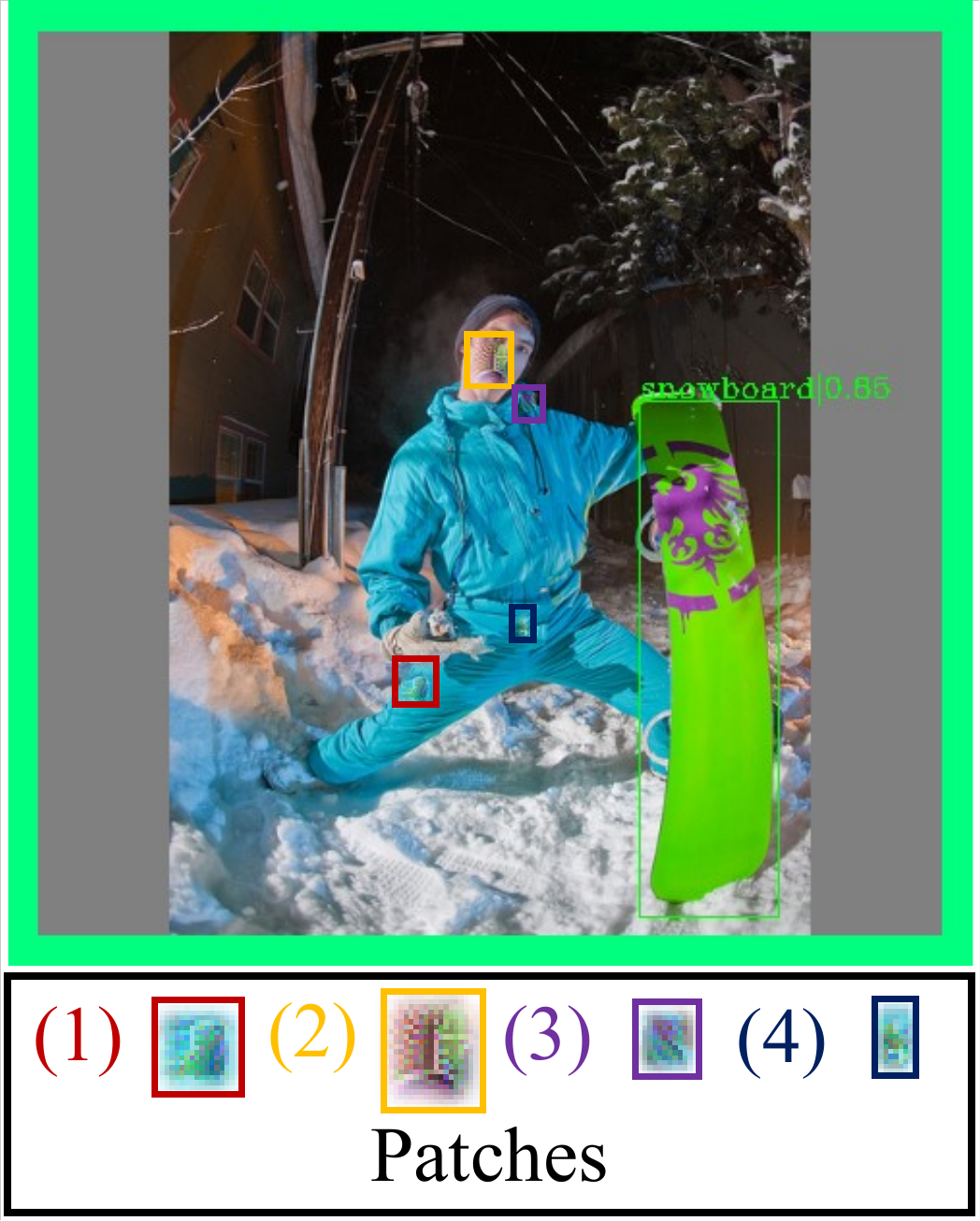}
}
\end{minipage}
\caption{
Adversarial examples of classification attack on Faster R-CNN~\cite{ren2015faster} on one image from COCO.
The detection results are shown in images. Some attacks eliminate the bounding boxes, while others mislead Faster R-CNN to detect the person to be a bird. 
The image with a red frame means that the image is recognized as an adversarial example by the adversarial patch detector, while a green frame means not recognized. 
We define $\text{P}_a$ as the probability of being adversarial as predicted by an adversarial patch detector, of which the probability threshold is 50\%. The detection results and $\text{P}_a$ of attack methods are:
(a)~original image, person = 100\%;
(b)~DPatch~\cite{liu2019dpatch}, nothing detected, $\text{P}_a$=90.28\%; 
(c)~AdvPatch~\cite{thys2019fooling}, nothing detected, $\text{P}_a$=66.84\%; 
(d)~$RP_2$-Sticker~\cite{eykholt2018physical}, nothing detected, $\text{P}_a$=65.56\%; 
(e)~2-Rects-H, bird = 96\% , $\text{P}_a$=71.55\%; 
(f)~4-Rects, bird = 76\%, $\text{P}_a$=60.03\%;
(g)~UPC-7P~\cite{huang2020universal}, nothing detected, $\text{P}_a$=57.29\%; and 
(h)~LDAP, nothing detected, $\text{P}_a$=24.06\%.
}
\label{fig:coco_untarget}
\end{figure}

\begin{table}[t!]
\centering  \scriptsize
\caption{
Performance of attack methods on classification attack on the COCO dataset.
ASR denotes Attack Success Rate;
``Area Thr'' denotes Area Threshold;
MAR denotes Mean Area Ratio; and
ADA denotes adversarial detection accuracy.
}
\label{tab:coco_untarget}
\setlength{\tabcolsep}{1.7mm}{
\begin{tabular}{c|c|cccc}
\hline
Detector       & Attack Method                 & \makecell[c]{ASR~(\%) \\ (Area Thr~(\%)) }     & \makecell[c]{MAR\\(\%)}  & \makecell[c]{LPIPS \\ ($\times 10^{-3}$) } & \makecell[c]{ADA\\(\%)} \\ \hline
\multirow{7}{*}{FRCN}
                            & DPatch~\cite{liu2019dpatch}   
                            & 0~(10)   & 96.70   & 101.3 & 97.2 \\ 
                            & AdvPatch~\cite{thys2019fooling}   
                            & 6.18~(10)   & 18.57    & 26.49 & 87.5\\ 
                            & $RP_2$-Sticker~\cite{eykholt2018physical}            
                            & 7.35~(10)  & 14.72    & 26.00 & 88.6 \\ 
                            & UPC-7P~\cite{huang2020universal}           
                            & 47.86~(10)  & 10.12    & 28.85 & 89.8 \\ 
                            & 2-Rects-H            
                            & 10.66~(10)  & 18.41    &  32.91 & 89.7 \\ 
                            & 4-Rects            
                            & 35.96~(10)  & 12.82    & 31.43 & 91.0 \\ 
                            & LDAP         
                            & \textbf{63.61}~(10)  & \textbf{9.36}  & \textbf{12.23}   & \textbf{73.1} \\ \hline
\multirow{7}{*}{MRCN}       
                            & DPatch~\cite{liu2019dpatch}   
                            & 0~(10)   & 97.42   & 101.7  & 96.6 \\ 
                            & AdvPatch~\cite{thys2019fooling}  
                            & 4.47~(10) & 21.01    &  30.73 & 88.9 \\ 
                            & $RP_2$-Sticker~\cite{eykholt2018physical}
                            & 20.42~(10)  & 16.58     &  28.75 & 91.2\\ 
                            & UPC-7P~\cite{huang2020universal}           
                            & 40.16~(10)  & 11.40    & 30.67 & 91.4 \\ 
                            & 2-Rects-H            
                            & 7.42~(10) & 19.85  & 35.51  & 91.6 \\ 
                            & 4-Rects            
                            & 20.97~(10)   & 15.81 & 35.79  & 92.4 \\ 
                            & LDAP         
                            & \textbf{60.93}~(10) & \textbf{9.70} &  \textbf{15.12}   & \textbf{71.1}   \\  \hline
\multirow{7}{*}{SSD}          
                            & DPatch~\cite{liu2019dpatch}   
                            & 2.68~(7)   & 39.07    & 47.3 & 96.5\\ 
                            & AdvPatch~\cite{thys2019fooling}  
                            & 59.83~(7)  & 7.17      & 13.13   & 78.4    \\ 
                            & $RP_2$-Sticker~\cite{eykholt2018physical}     
                            & 45.87~(7)  & 8.89       &  18.20   & 86.8   \\ 
                            & UPC-7P~\cite{huang2020universal}           
                            & 47.11~(7)  & 8.61  & 26.0 & 92.0 \\ 
                            & 2-Rects-H            
                            & 44.70~(7) & 9.02   &  20.42   & 87.6 \\ 
                            & 4-Rects            
                            & 41.26~(7)  & 9.59   &   25.93 & 93.4  \\ 
                            & LDAP         
                            & \textbf{63.06}~(7) & \textbf{6.39}       &  \textbf{9.33}  & \textbf{76.6}  \\ \hline
\multirow{7}{*}{YOLOv3}       
                            & DPatch~\cite{liu2019dpatch}   
                            & 1.92~(6)   & 43.89    & 52.7 & 94.3\\
                            & AdvPatch~\cite{thys2019fooling}  
                            & 46.90~(6) & 8.17      &   13.23   & 75.0   \\ 
                            & $RP_2$-Sticker~\cite{eykholt2018physical}            
                            & 48.07~(6) & 8.42        &  16.10    & 78.0  \\
                            & UPC-7P~\cite{huang2020universal}           
                            & 63.06~(6)  & 5.74    & 19.83 & 85.3 \\  
                            & 2-Rects-H            
                            & 52.13~(6) & 7.51      &  16.61    & 78.6  \\ 
                            & 4-Rects            
                            & 55.29~(6) & 6.96        &  20.05  & 83.2  \\ 
                            & LDAP         
                            & \textbf{63.75}~(6) & \textbf{5.61}   & \textbf{8.98} & \textbf{62.5} \\ \hline
\end{tabular}
}
\end{table}

\begin{figure}[!ht]
\centering
\includegraphics[width=0.7\linewidth]{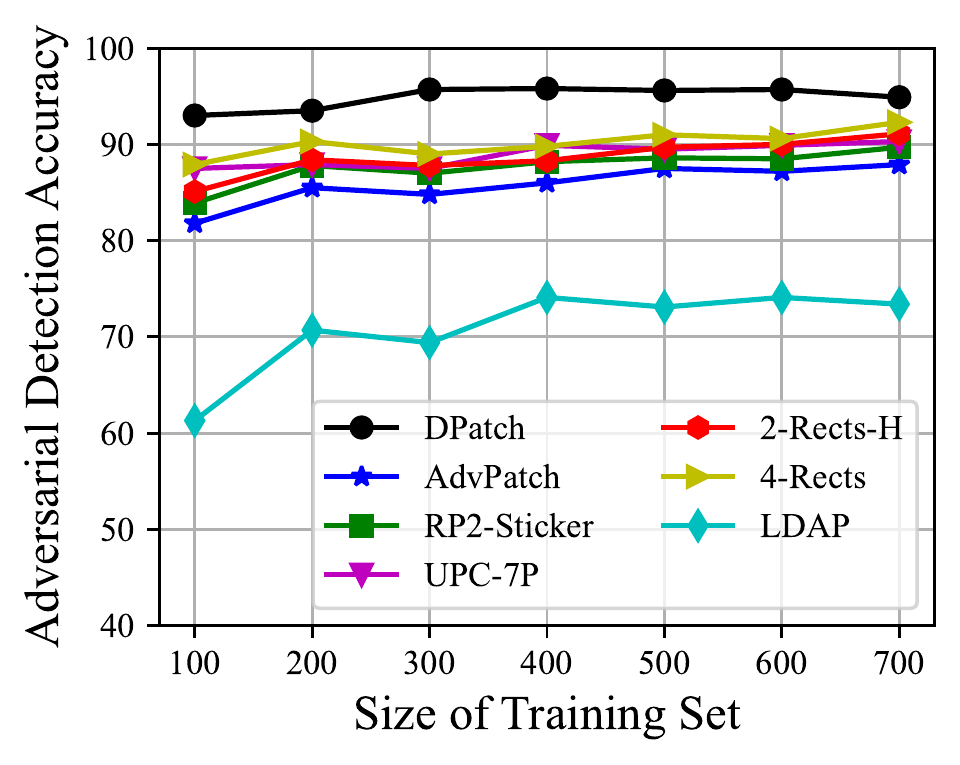}
\caption{
The relationship of training set size and adversarial detection accuracy for different attack methods on classification attack on Faster R-CNN.}
\label{fig:train_set_influence}
\end{figure}

Demonstrations of different attack methods on a classification-attack task on Faster R-CNN are shown in Fig.~\ref{fig:coco_untarget}. 
We find that: \textbf{(i)}~all attack methods successfully attack the Faster R-CNN, while only LDAP passes the detection of the adversarial patch detector;
\textbf{(ii)}~as for region and texture consistency, we see that LDAP takes the least patch area, while the texture of patches generated by LDAP is the most consistent with that of the original image (in contrast, DPatch takes the largest patch);
\textbf{(iii)}~although the number of rectangle primitives is set to 10 in LDAP, the number of patches in the adversarial example of LDAP is much less than 10, indicating that the heights and widths of some rectangle primitives are decreased to zero during optimization; and 
\textbf{(iv)}~searched patches of LDAP do not connect to each other to form more complex geometric shapes, suggesting that more dispersed patches, not more complex shapes, may help reduce patch area.

Table~\ref{tab:coco_untarget} illustrates the attack performance and detectability of LDAP and comparison methods, and provides the following insights:
\textbf{(i)}~for all four detection models, LDAP gets the highest ASR under the same region area threshold, the smallest patches, the lowest LPIPS, and the lowest adversarial detection accuracy, outperforming other attack methods by a large margin; note that the lowest value of adversarial detection accuracy is 50\%, meaning random guessing, reducing adversarial detection accuracy from 78.4\%~(AdvPatch~\cite{thys2019fooling}) to 76.6\%, although still a big improvement when we attack SSD;
\textbf{(ii)}~for all attack methods, two-stage detectors need more patch area than one-stage detectors, which indicates they are more difficult to attack (this phenomenon is also found on other attack tasks);
\textbf{(iii)}~one interesting finding is that, among all comparison methods, a greater number of patches usually results in less area of patches, but for SSD this conclusion does not hold. This phenomenon is also found on other attack tasks.
We conjecture that this is because SSD is more sensitive to the middle region during attack, which is exactly where AdvPatch~\cite{thys2019fooling} attacks; 
\textbf{(iv)}~another finding is that, although UPC-7P~\cite{huang2020universal} takes the smallest patch area among the comparison methods on detectors except SSD, it still gets high LPIPS and adversarial detection accuracy, which means that a small region area does not sufficiently represent low detectability; and
\textbf{(v)}~DPatch~\cite{liu2019dpatch} takes the largest patch area on all models. We attribute this phenomenon to the upper-left corner patch position of DPatch. If the patch does not increase its area to reach the target object, the receptive field of the target object in detectors cannot be influenced by the patch, leading to a failed attack. 
Considering that the target object is usually not in the upper-left corner of the image, the patch must be large enough to reach the target object, resulting in a large patch. 
Due to the large patch, the ADA of DPatch is also the highest.

\begin{figure}[t!]
\centering 
\begin{minipage}[]{.19\linewidth} \scriptsize \centering
\subfloat[]{
\includegraphics[width=\linewidth]{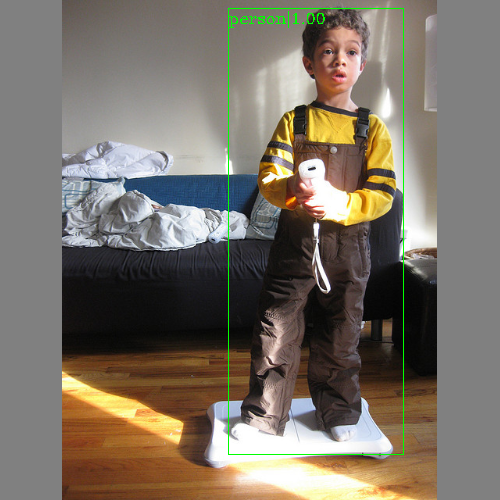}
}
\end{minipage}
\begin{minipage}[]{.6\linewidth} \scriptsize \centering
\subfloat[]{
\includegraphics[width=.32\linewidth]{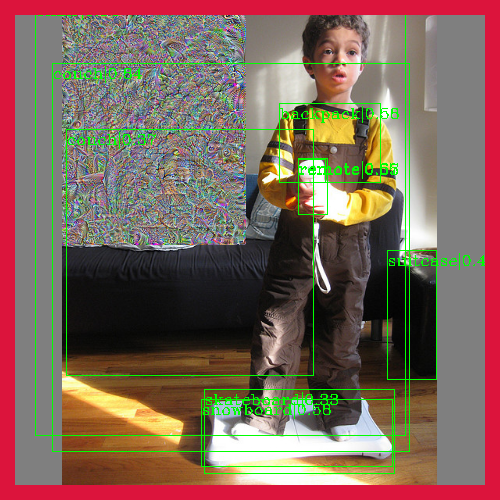}
}
\subfloat[]{
\includegraphics[width=.32\linewidth]{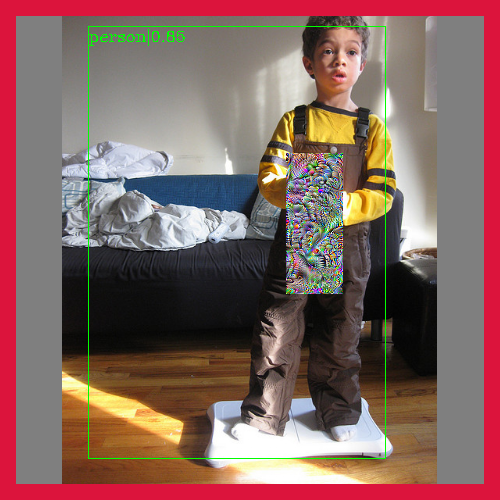}
}
\subfloat[]{
\includegraphics[width=.32\linewidth]{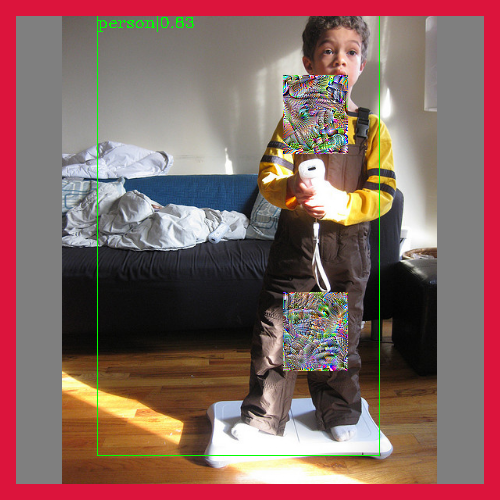}
}
\\
\subfloat[]{
\includegraphics[width=.32\linewidth]{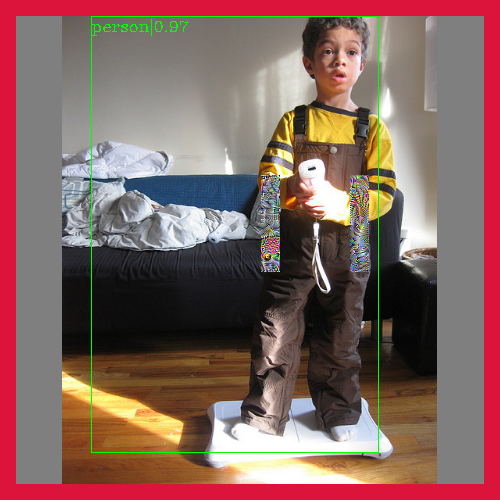}
}
\subfloat[]{
\includegraphics[width=.32\linewidth]{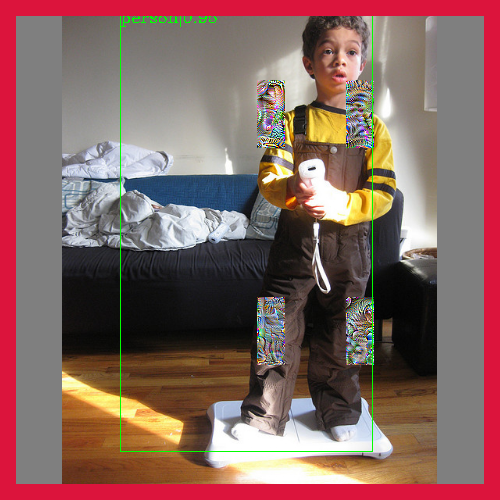}
}
\subfloat[]{
\includegraphics[width=.32\linewidth]{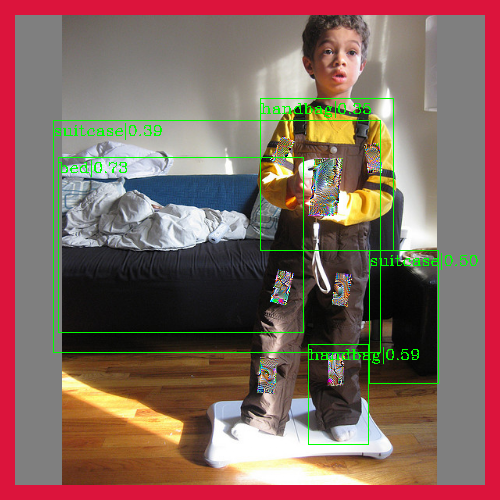}
}
\end{minipage}
\begin{minipage}[]{.19\linewidth} \scriptsize \centering
\subfloat[]{
\includegraphics[width=\linewidth]{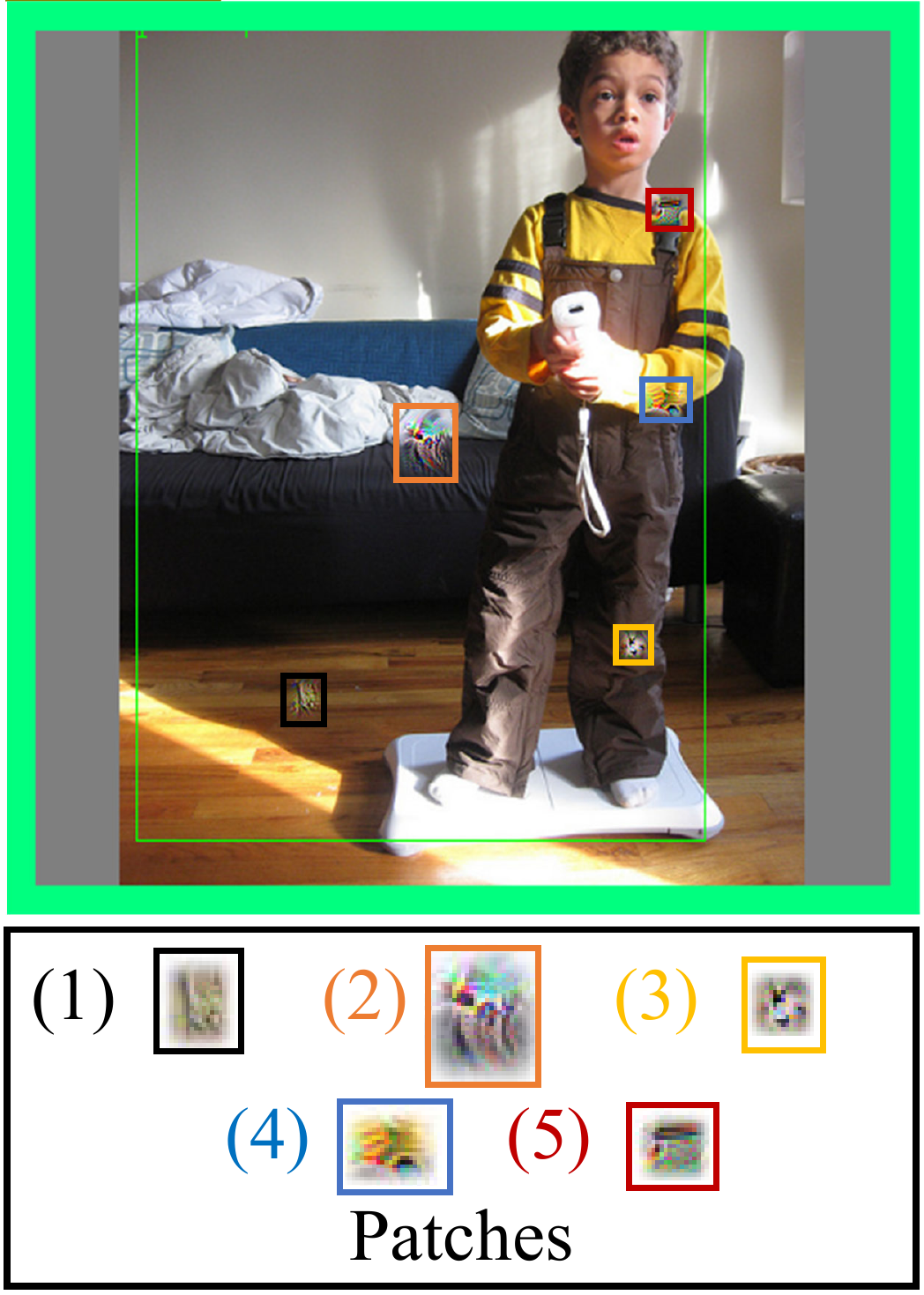}
}
\end{minipage}
\caption{
Adversarial examples of localization attack on Mask R-CNN~\cite{he2017mask} on one image from COCO. The detection results are also shown in images. 
Some attacks hide the person, while others shift the bounding boxes.
The maximum IoU between the predicted bounding boxes and the ground-truth bounding box and probabilities predicted by corresponding adversarial patch detectors are:
(a)~original image, person=100\%, IoU=95.05\%;
(b)~DPatch~\cite{liu2019dpatch}, no person detected, $\text{P}_a$=96.38\%; 
(c)~AdvPatch~\cite{thys2019fooling}, IoU = 47.03\%, $\text{P}_a$=90.41\%; 
(d)~$RP_2$-Sticker~\cite{eykholt2018physical}, IoU = 46.65\%, $\text{P}_a$=89.45\%; 
(e)~2-Rects-H, IoU = 46.65\%, $\text{P}_a$=80.65\%; 
(f)~4-Rects, IoU = 49.98\%, $\text{P}_a$=85.92\%; 
(g)~UPC-7P~\cite{huang2020universal}, no person detected, $\text{P}_a$=77.36\%; and
(h)~LDAP, IoU = 44.20\%, $\text{P}_a$=38.29\%.
}
\label{fig:coco_regress}
\end{figure}

\begin{table}[!ht]
\centering \scriptsize
\caption{
Performance of attack methods on localization attack on COCO dataset.}
\setlength{\tabcolsep}{1.7mm}{
\begin{tabular}{c|c|cccc}
\hline
Detector       & Attack Method                 & \makecell[c]{ASR~(\%) \\ (Area Thr~(\%)) }     & \makecell[c]{MAR\\(\%)}  & \makecell[c]{LPIPS \\ ($\times 10^{-3}$) } & \makecell[c]{ADA\\(\%)} \\ \hline
\multirow{7}{*}{FRCN} 
                            & DPatch~\cite{liu2019dpatch}   
                            & 0.20~(14)   & 122.60    & 127.25 & 95.5\\
                            & AdvPatch~\cite{thys2019fooling}  
                            & 9.90~(14)     & 26.14  &    35.88   & 88.5        \\ 
                            & $RP_2$-Sticker~\cite{eykholt2018physical}            
                            & 21.11~(14)     & 21.67  &   35.49    & 90.2     \\ 
                            & UPC-7P~\cite{huang2020universal}           
                            & 50.68~(14)  & 14.22  & 33.83 & 90.9 \\  
                            & 2-Rects-H            
                            & 10.59~(14)   & 24.81   &  42.65    & 89.4   \\ 
                            & 4-Rects            
                            & 15.61~(14)     & 23.53  &  46.79   & 92.5     \\ 
                            & LDAP         
                            & \textbf{64.16}~(14)   & \textbf{13.85}    &  \textbf{23.48}      & \textbf{74.5}   \\ \hline
\multirow{7}{*}{MRCN}   
                            & DPatch~\cite{liu2019dpatch}   
                            & 0.13~(15)   & 123.40    & 128.95 & 96.8\\
                            & AdvPatch~\cite{thys2019fooling}  
                            & 11.96~(15)   & 26.87   &   36.59  & 87.9    \\ 
                            & $RP_2$-Sticker~\cite{eykholt2018physical}            
                            & 24.34~(15)   & 21.83   &   35.82    & 90.8  \\
                            & UPC-7P~\cite{huang2020universal}           
                            & 54.81~(15)   & 14.55   & 34.19 & 90.6 \\   
                            & 2-Rects-H            
                            & 11.82~(15)   & 25.70  &   43.80    & 91.0    \\ 
                            & 4-Rects            
                            & 14.16~(15)   & 24.80   &   48.55   & 92.4    \\ 
                            & LDAP         
                            & \textbf{64.37}~(15)   & \textbf{14.11}   &   \textbf{25.87}   & \textbf{71.7}    \\  \hline
\multirow{7}{*}{SSD}          
                            & DPatch~\cite{liu2019dpatch}   
                            & 0.89~(9)   & 53.57    & 62.02 & 97.9\\
                            & AdvPatch~\cite{thys2019fooling}  
                            & 49.31~(9)  & 9.85    &   16.42  & 80.7     \\ 
                            & $RP_2$-Sticker~\cite{eykholt2018physical}            
                            & 39.20~(9)  & 11.10     &  21.74  & 88.6   \\
                            & UPC-7P~\cite{huang2020universal}           
                            & 46.14~(9)  & 9.70   & 28.9 & 93.1 \\  
                            & 2-Rects-H            
                            & 35.55~(9)  & 11.84   &  24.83   & 89.9     \\ 
                            & 4-Rects            
                            & 27.44~(9)  & 12.41   &   31.38  & 94.0     \\
                            & LDAP         
                            & \textbf{59.07}~(9)  & \textbf{8.83}   &  \textbf{12.74}  & \textbf{75.2}    \\ \hline
\multirow{7}{*}{YOLOv3}       
                            & DPatch~\cite{liu2019dpatch}   
                            & 0.96~(10)   & 72.74    & 78.83 & 94.3\\
                            & AdvPatch~\cite{thys2019fooling}  
                            & 26.96~(9)  & 13.74     &   20.37  & 82.0    \\ 
                            & $RP_2$-Sticker~\cite{eykholt2018physical}            
                            & 30.33~(9)  & 12.99    &   23.14  & 84.9     \\ 
                            & UPC-7P~\cite{huang2020universal}           
                            & 44.42~(9)  & 9.90     & 27.43     & 90.1 \\ 
                            & 2-Rects-H            
                            & 26.89~(9)  & 13.20    &  25.56   & 84.1     \\ 
                            & 4-Rects            
                            & 24.00~(9)  & 13.28     &   30.76  & 89.2   \\
                            & LDAP         
                            & \textbf{63.61}~(9)  & \textbf{8.31}   &  \textbf{11.94}   & \textbf{68.8}    \\ \hline

\end{tabular}
}
\label{tab:coco_regress}
\end{table}

In addition to the comparison of adversarial detection accuracy with the fixed size of the training set, we also test the relationship of adversarial detection accuracy and the size of the training set. The comparisons are shown in Fig.~\ref{fig:train_set_influence}. We find that the curve of LDAP is much lower than those of comparison methods. This indicates that the detectability of our attack is lower than that of other methods at various sizes of the training set.

\subsubsection{Localization Attack}
\label{sec:Localization_Attack}
Localization attack means the adversary should mislead the detector to predict a bounding box with undesirable shape or position on the target object.
In localization attack, we study the horizontal bounding-box shift attack. 
The attack success criterion is set as the maximum IoU of the predicted bounding boxes within the correct category is lower than 0.5. 

Demonstrations of localization attack on Mask R-CNN~\cite{he2017mask} are shown in Fig~\ref{fig:coco_regress}. 
We find that the successful attack can be achieved not only by moving the bounding box, but also by vanishing the bounding boxes within the correct category, such as DPatch and UPC-7P.
Another finding is that, for the horizontal bounding-box shift attack, horizontally-distributed patches such as 2-Rects-H have natural advantages that enable their patches to be smaller. 
Our LDAP also learns to arrange the patches horizontally. As for detectability, LDAP is the only one that escapes the detection of the adversarial patch detector.

Detailed attack results are shown in Table~\ref{tab:coco_regress}. We find that \textbf{(i)}~LDAP still achieves the best performance on all detectors; and 
\textbf{(ii)}~compared with the classification attack task, the localization attack task requires more patch area.

\subsection{Experiments on $\text{D}^2$-City}
\label{sec:Experiments_on_dd}
We investigate the threat of LDAP in an autonomous driving scenario on the $\text{D}^2$-City dataset. Considering the poor performance of DPatch~\cite{liu2019dpatch}, we do not include it in the comparison on $\text{D}^2$-City. Besides, UPC-7P is specifically designed for the attack ``person'' category, so we do not make comparisons with it, either.
In $\text{D}^2$-City, two different threatening attack cases are studied, including \emph{vanishing a car ahead} and \emph{shifting a car to the wrong lane}.

\subsubsection{Vanishing a car ahead}
This attack is dangerous, since it may directly lead the autonomous car to hit the car in front of it. The attack is essentially an application of a classification attack.
The attack success criterion is set to be the same as the criterion of a classification attack on COCO, which is that there is no bounding box with a correct category that overlaps the bounding box of the target object.

\begin{figure}[t!]
\centering 
\begin{minipage}[]{.24\linewidth} \scriptsize \centering
\subfloat[]{
\includegraphics[width=\linewidth]{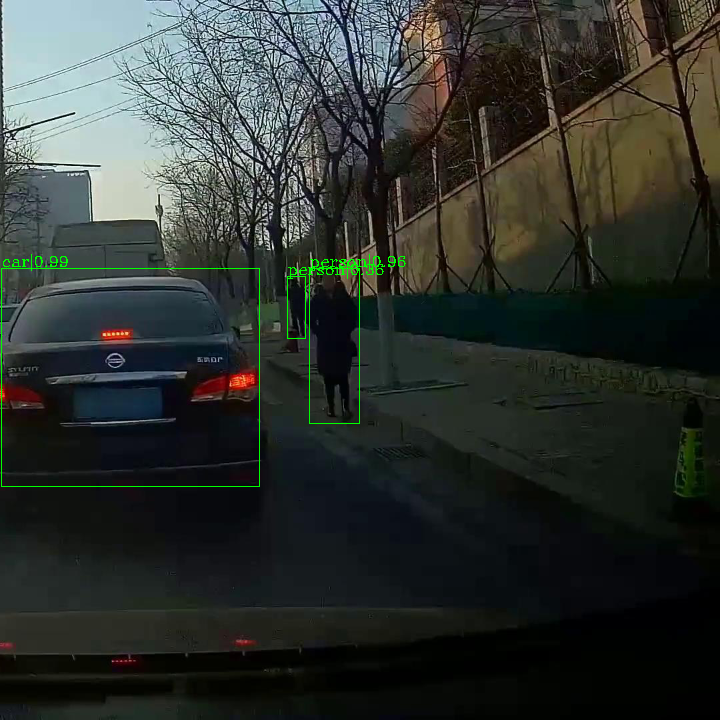}
}
\end{minipage}
\begin{minipage}[]{.5\linewidth} \scriptsize \centering
\subfloat[]{
\includegraphics[width=.49\linewidth]{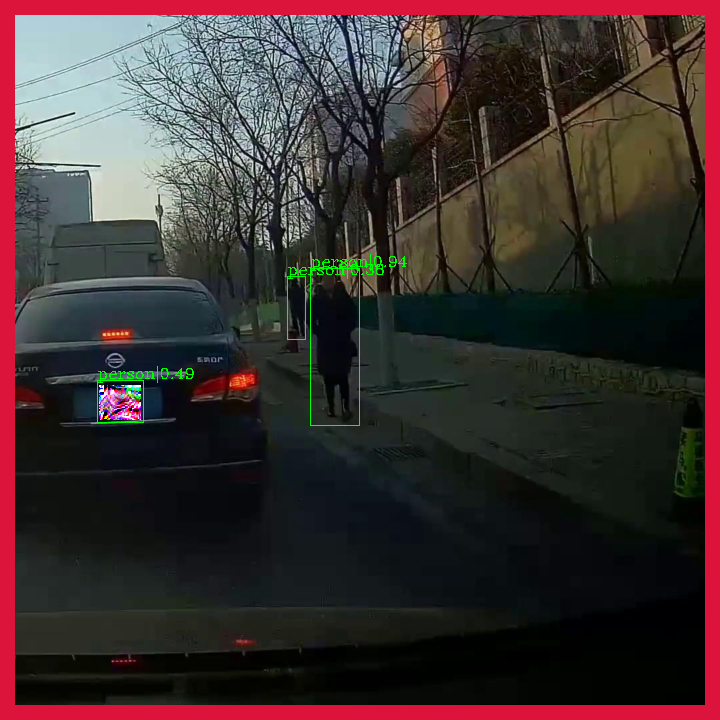}
}
\subfloat[]{
\includegraphics[width=.49\linewidth]{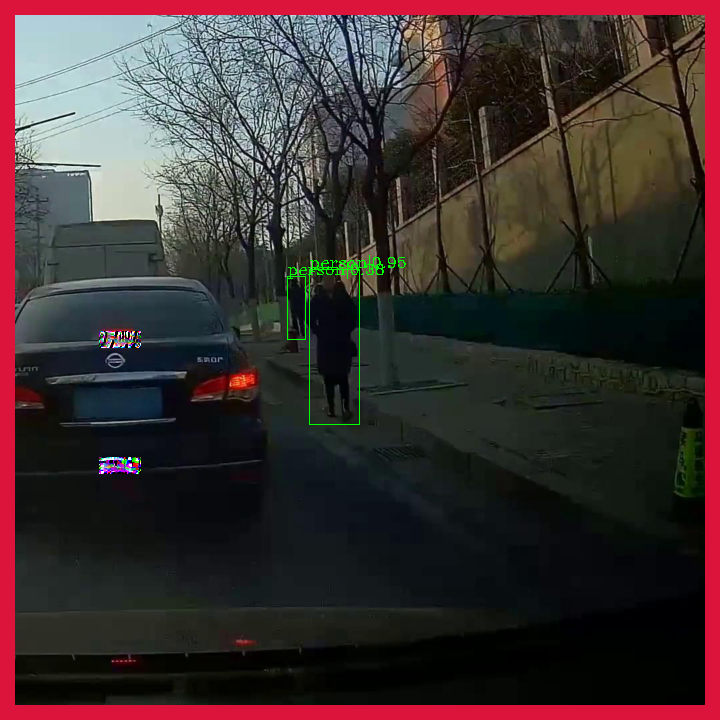}
}
\\
\subfloat[]{
\includegraphics[width=.49\linewidth]{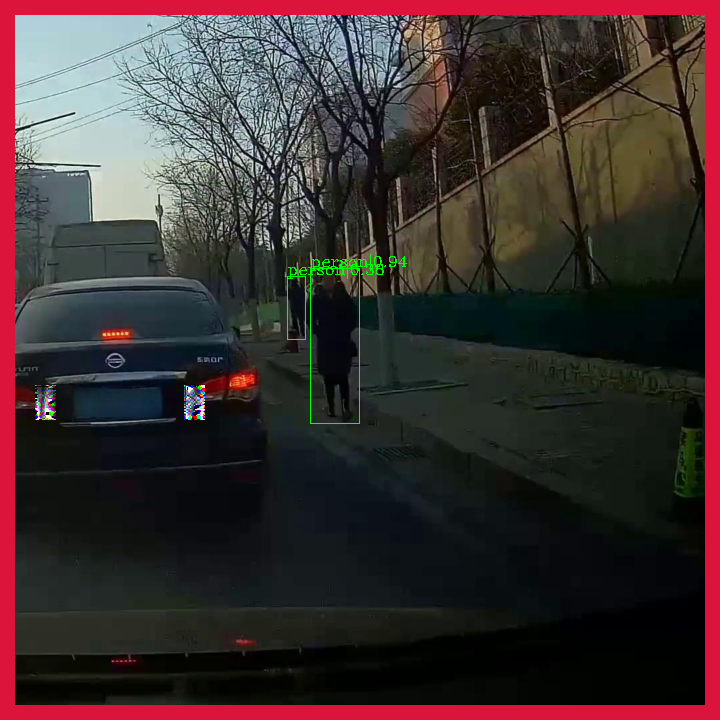}
}
\subfloat[]{
\includegraphics[width=.49\linewidth]{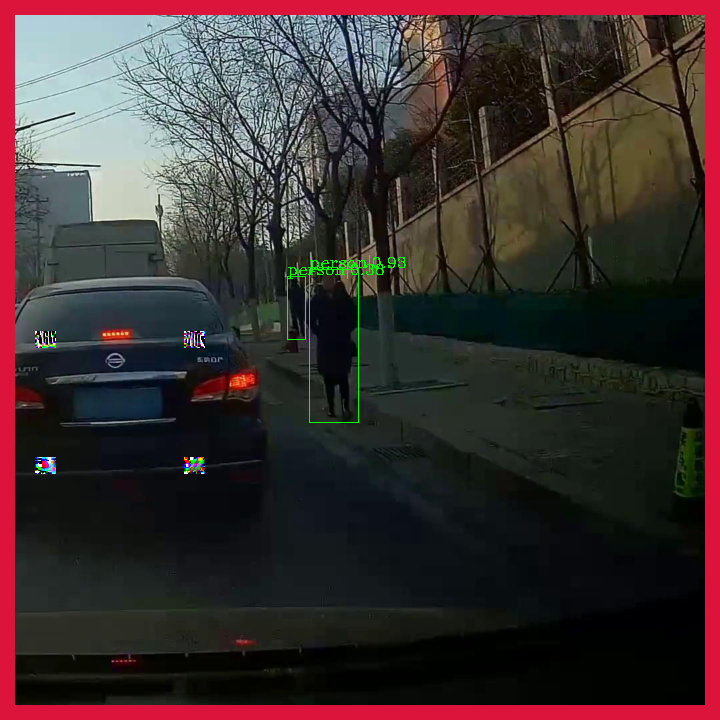}
}
\end{minipage}
\begin{minipage}[]{.24\linewidth} \scriptsize \centering
\subfloat[]{
\includegraphics[width=\linewidth]{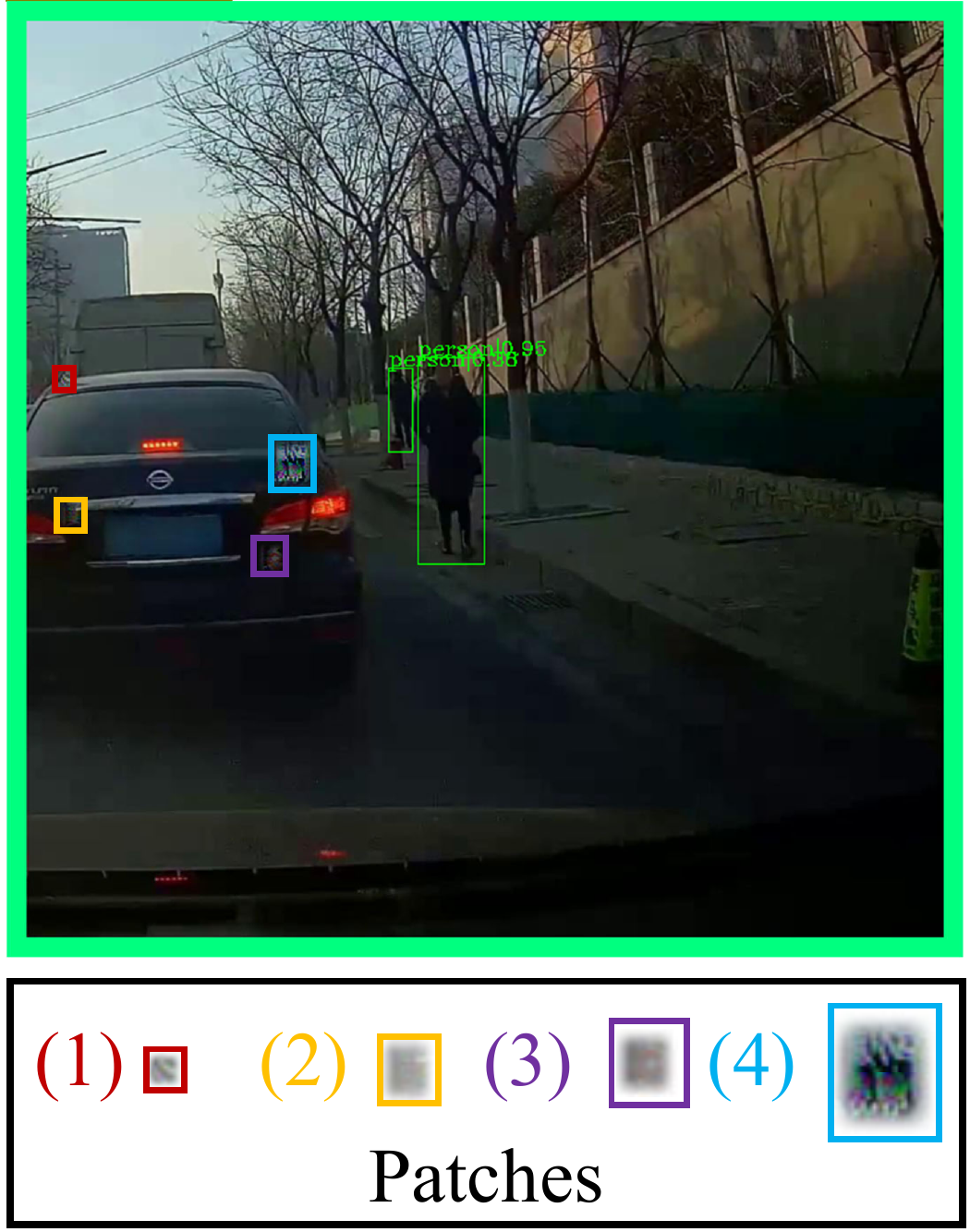}
}
\end{minipage}
\caption{
Adversarial examples of \emph{vanishing a car ahead} attack task on SSD~\cite{liu2016ssd} on one image from $\text{D}^2$-City. 
The detection results and probabilities predicted by corresponding adversarial patch detectors are:
(a)~original image, car = 99\%;
(b)~AdvPatch~\cite{thys2019fooling}, person = 49\%, $\text{P}_a$=84.19\%; 
(c)~$RP_2$-Sticker~\cite{eykholt2018physical}, nothing detected, $\text{P}_a$=77.79\%; 
(d)~2-Rects-H, nothing detected, $\text{P}_a$=74.43\%;  
(e)~4-Rects, nothing detected, $\text{P}_a$=64.70\%; and 
(f)~LDAP, nothing detected, $\text{P}_a$=36.66\%.
}
\label{fig:didi_untarget}
\end{figure}

The demonstration of this attack task is shown in Fig.~\ref{fig:didi_untarget}. A first glance shows that the patches of LDAP are nearly imperceptible compared with those of other attack methods. Besides, the adversarial example of LDAP is the only one that passes the adversarial patch detector.

Comparisons are shown in Table~\ref{tab:didi_untarget}. 
The findings are similar to experiments in COCO, and our LDAP still gets the best performance on all object detectors.
One interesting finding is that the attack on cars in the $\text{D}^2$-City dataset requires less patch area than the attack on persons in the COCO dataset. We attribute this to the low image quality of driving-record video, which makes cars in $\text{D}^2$-City much easier to attack than the persons in COCO, resulting in less MAR.

\subsubsection{Shifting a car to the wrong lane}
This attack may influence the lane-change and speed-control decisions of an autonomous car, causing a collision. Clearly, it is an important attack case to study. This attack is essentially an application of the bounding-box shift attack.
We set the attack success criterion to be the same as the criterion for a localization attack on COCO, which is that the max IoU between bounding boxes with the correct category and target bounding box is lower than 0.5.

\begin{figure}[t!]
\centering 
\begin{minipage}[]{.24\linewidth} \scriptsize \centering
\subfloat[]{
\includegraphics[width=\linewidth]{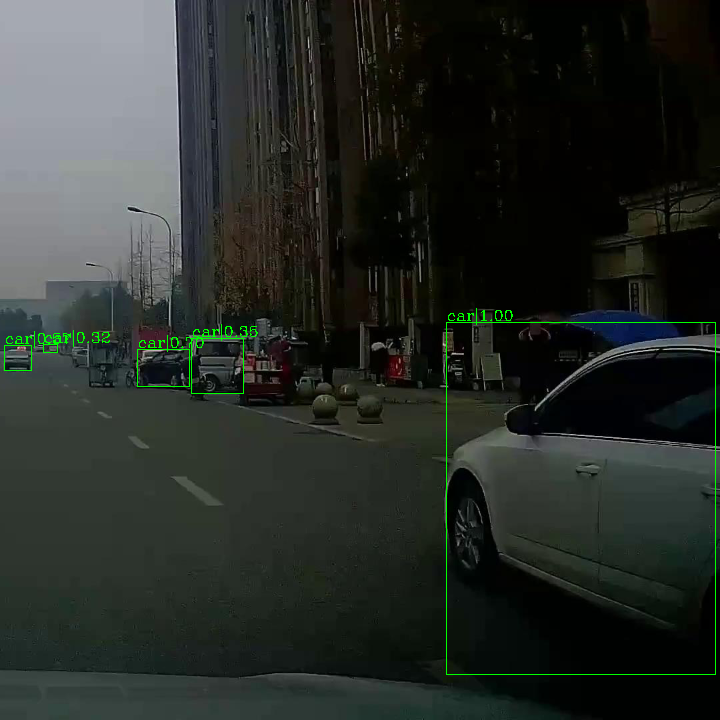}
}
\end{minipage}
\begin{minipage}[]{.5\linewidth} \scriptsize \centering
\subfloat[]{
\includegraphics[width=.49\linewidth]{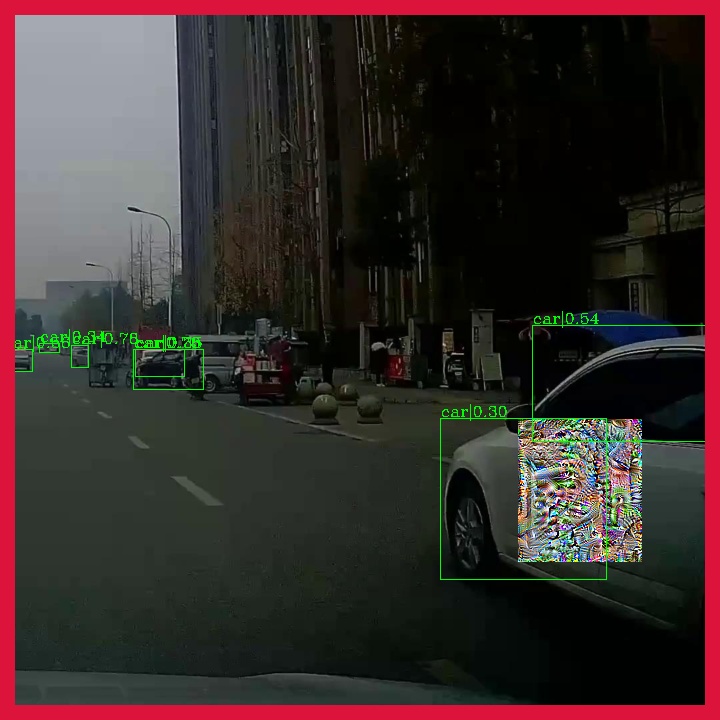}
}
\subfloat[]{
\includegraphics[width=.49\linewidth]{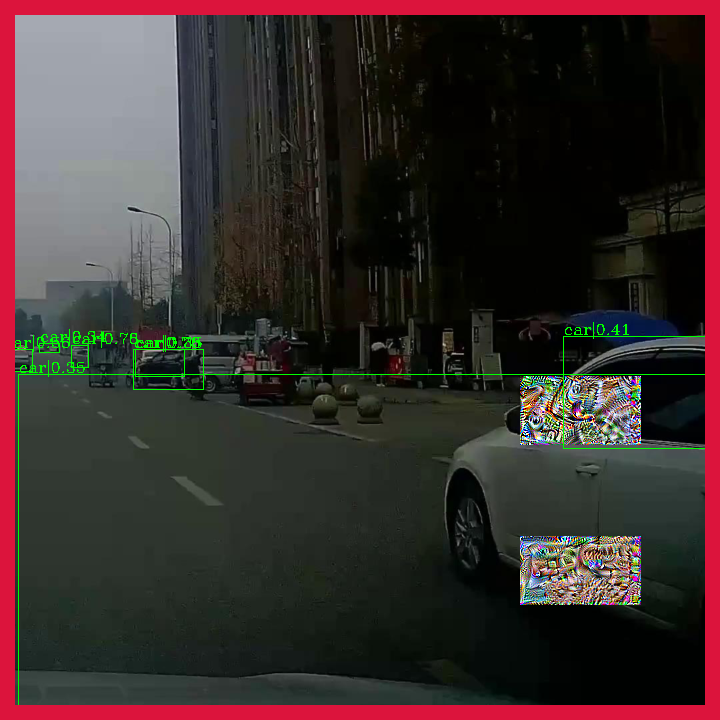}
}
\\
\subfloat[]{
\includegraphics[width=.49\linewidth]{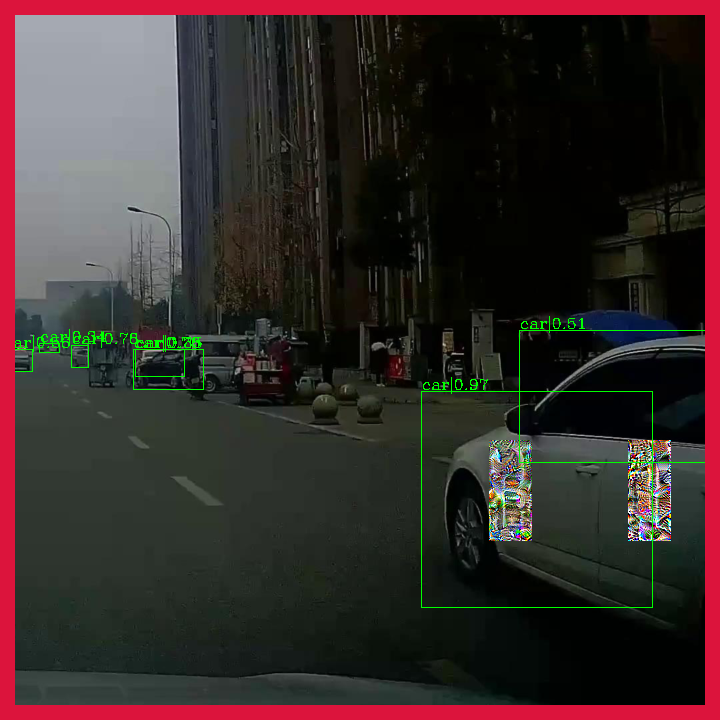}
}
\subfloat[]{
\includegraphics[width=.49\linewidth]{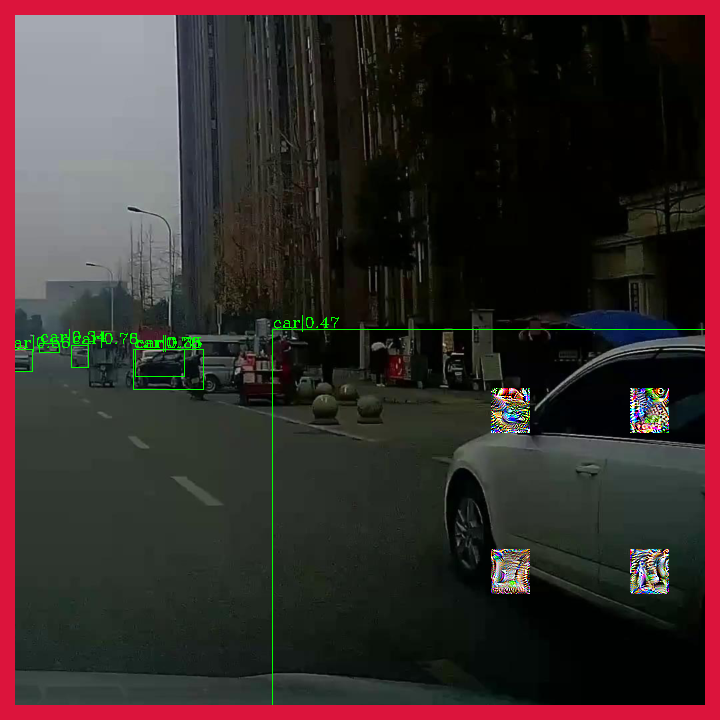}
}
\end{minipage}
\begin{minipage}[]{.24\linewidth} \scriptsize \centering
\subfloat[]{
\includegraphics[width=\linewidth]{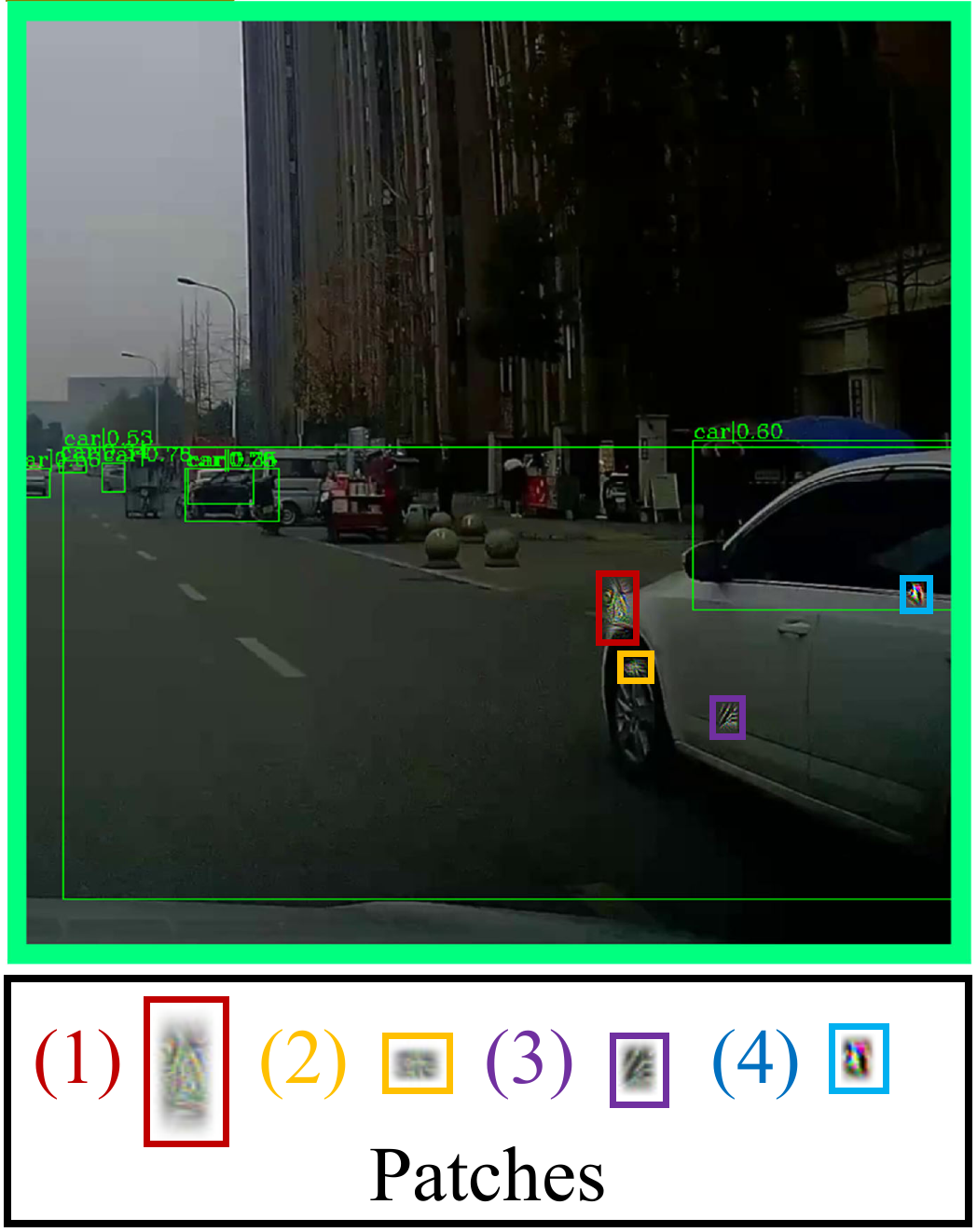}
}
\end{minipage}
\caption{
Adversarial examples of \emph{shift a car to the wrong lane} attack task on YOLOv3~\cite{redmon2018yolov3} on one image from $\text{D}^2$-City. 
The maximum IoU and the probabilities predicted by the adversarial patch detectors are:
(a)~original image, car = 100\%, IoU = 88.19\%;
(b)~AdvPatch~\cite{thys2019fooling}, IoU = 29.61\%, $\text{P}_a$=90.78\%;
(c)~$RP_2$-Sticker~\cite{eykholt2018physical}, IoU = 30.25\%, $\text{P}_a$=91.91\%; 
(d);~2-Rects-H, IoU = 48.35\%, $\text{P}_a$=79.54\%; 
(e)~4-Rects, IoU = 49.67\%, $\text{P}_a$=70.86\%; and
(f)~LDAP, IoU = 38.53\%, $\text{P}_a$=29.06\%.
}
\label{fig:didi_regress}
\end{figure}

Examples of this attack are shown in Fig.~\ref{fig:didi_regress}. We see that our LDAP still utilizes the smallest patch area and still passes the detection of the adversarial patch detector. 
Furthermore, these images demonstrate different attack success situations in localization attack, such as by reducing the size of bounding boxes or by moving the boundary of bounding boxes to one side. 
Comparison results are shown in Table~\ref{tab:didi_regress}. 
We find that the LDAP still gets the best performance among all detectors.

\begin{table}[t!] 
\scriptsize
\centering
\caption{
\small
Performance of attacks on \emph{vanishing a car ahead} attack task. ADA for adversarial detection accuracy.}
\setlength{\tabcolsep}{1.7mm}{
\begin{tabular}{c|c|cccc}
\hline
Detector       & Attack Method                 & \makecell[c]{ASR~(\%) \\ (Area Thr~(\%)) }     & \makecell[c]{MAR\\(\%)}  & \makecell[c]{LPIPS \\ ($\times 10^{-3}$) } & \makecell[c]{ADA\\(\%)} \\ \hline
\multirow{5}{*}{FRCN} & AdvPatch~\cite{thys2019fooling}  
                              & 28.74~(5)  & 17.42       & 66.68    & 90.6    \\ 
                              & $RP_2$-Sticker~\cite{eykholt2018physical} 
                              & 41.05~(5)   & 15.23      & 67.08   & 92.5     \\ 
                              & 2-Rects-H            
                              & 44.98~(5)  & 12.26       & 67.02    & 92.3    \\ 
                              & 4-Rects            
                              & 56.93~(5)  & 8.37         & 64.40  & 91.6      \\ 
                              & LDAP         
                              & \textbf{63.50}~(5)  & \textbf{4.67}       & \textbf{49.81}  & \textbf{79.6}     \\ \hline
\multirow{5}{*}{MRCN}   & AdvPatch~\cite{thys2019fooling}  
                              & 34.21~(5)  & 16.85     & 65.69     & 90.6       \\ 
                              & $RP_2$-Sticker~\cite{eykholt2018physical}            
                              & 41.42~(5)  & 14.93     & 65.35  & 90.1     \\ 
                              & 2-Rects-H            
                              & 45.52~(5)  & 12.42      & 67.00   & 92.1     \\ 
                              & 4-Rects            
                              & 57.93~(5)  & 8.54     & 64.49  & 91.1   \\ 
                              & LDAP         
                              & \textbf{62.22}~(5)  & \textbf{4.68}      & \textbf{49.83}   & \textbf{79.0}     \\  \hline
\multirow{5}{*}{SSD}          & AdvPatch~\cite{thys2019fooling}  
                              & 60.40~(4)  & 6.15      & 47.73   & 82.0    \\ 
                              & $RP_2$-Sticker~\cite{eykholt2018physical}            
                              & 60.58~(4)  & 6.08      & 50.53   & 87.3    \\ 
                              & 2-Rects-H            
                              & 60.40~(4)  & 5.93     & 56.17   & 87.2     \\ 
                              & 4-Rects            
                              & 66.78~(4)  & 4.75     & 58.13   & 91.5     \\
                              & LDAP         
                              & \textbf{69.61}~(4)  & \textbf{3.88}     & \textbf{5.94}   & \textbf{71.0}    \\ \hline
\multirow{5}{*}{YOLOv3}       & AdvPatch~\cite{thys2019fooling}  
                              & 38.86~(3)  & 8.97     &  48.47  & 84.8     \\ 
                              & $RP_2$-Sticker~\cite{eykholt2018physical}            
                              & 46.53~(3)  & 7.81    &  49.78  & 87.0     \\ 
                              & 2-Rects-H            
                              & 44.43~(3)  & 7.09      & 57.38   & 87.6     \\ 
                              & 4-Rects            
                              & 54.56~(3)  & 5.57       & 58.73  & 90.7    \\
                              & LDAP         
                              & \textbf{63.50}~(3)  & \textbf{2.81}     & \textbf{6.70}     & \textbf{65.1}     \\ \hline
\end{tabular}
}
\label{tab:didi_untarget}
\end{table}

\subsection{Ablation Study}
\label{sec:Ablation_Study}
We perform an ablation study on the rectangle primitive number $N$, soft-attack strategy, mask layer search and texture constraint.

\begin{table}[t] 
\scriptsize
\centering
\caption{
Performance of attacks on \emph{shift a car to the wrong lane} attack task.}
\setlength{\tabcolsep}{1.7mm}{
\begin{tabular}{c|c|cccc}
\hline
Detector       & Attack Method                 & \makecell[c]{ASR~(\%) \\ (Area Thr~(\%)) }     & \makecell[c]{MAR\\(\%)}  & \makecell[c]{LPIPS \\ ($\times 10^{-3}$) } & \makecell[c]{ADA\\(\%)} \\ \hline
\multirow{5}{*}{FRCN} & AdvPatch~\cite{thys2019fooling}  
                              &  25.72~(9)   & 12.98  & 63.69      & 88.1         \\ 
                              & $RP_2$-Sticker~\cite{eykholt2018physical}            
                              &  31.84~(9)   & 12.31    & 67.33   & 93.5     \\ 
                              & 2-Rects-H            
                              &  27.55~(9)  & 13.2     & 71.49    & 94.3    \\ 
                              & 4-Rects            
                              &  40.96~(9)   & 10.56    & 72.67  & 95.3     \\ 
                              & LDAP         
                              &  \textbf{68.97}~(9)  & \textbf{8.01}      & \textbf{53.25} & \textbf{74.0}     \\ \hline
\multirow{5}{*}{MRCN}   & AdvPatch~\cite{thys2019fooling}  
                              &  20.16~(8) & 13.1   & 63.87     & 89.0       \\ 
                              & $RP_2$-Sticker~\cite{eykholt2018physical}            
                              &  21.25~(8) & 12.97    & 68.45  & 93.3     \\ 
                              & 2-Rects-H            
                              &  21.35~(8) & 13.16  & 71.45   & 93.7    \\ 
                              & 4-Rects            
                              &  28.64~(8) & 10.64   & 72.75  & 95.3     \\ 
                              & LDAP         
                              &  \textbf{58.66}~(8)  & \textbf{7.93}    & \textbf{57.05}  & \textbf{74.3}    \\  \hline
\multirow{5}{*}{SSD}          & AdvPatch~\cite{thys2019fooling}  
                              &  30.47~(8) & 13.87   & 62.82   & 87.8    \\ 
                              & $RP_2$-Sticker~\cite{eykholt2018physical}            
                              &  27.64~(8) & 15.32   & 69.46  & 94.1    \\ 
                              & 2-Rects-H            
                              &  20.89~(8) & 17.35    & 76.79   & 93.1    \\ 
                              & 4-Rects            
                              &  25.27~(8)  & 15.90   & 80.11   & 95.0    \\
                              & LDAP         
                              &  \textbf{62.68}~(8) & \textbf{7.91}   & \textbf{15.07}  & \textbf{80.0}    \\ \hline
\multirow{5}{*}{YOLOv3}       & AdvPatch~\cite{thys2019fooling}  
                              & 53.92~(12)  & 14.30      &  63.07  & 84.7    \\ 
                              & $RP_2$-Sticker~\cite{eykholt2018physical}            
                              & 57.75~(12) & 12.92      &  63.79  & 87.1     \\ 
                              & 2-Rects-H            
                              & 55.83~(12) & 14.19      &  71.51  & 90.1     \\ 
                              & 4-Rects            
                              & 60.12~(12)  & 12.61     & 72.71  & 88.7  \\
                              & LDAP         
                              & \textbf{63.50}~(12) & \textbf{11.66}      & \textbf{10.64}  & \textbf{62.5}     \\ \hline
\end{tabular}
}
\label{tab:didi_regress}
\end{table}

\subsubsection{Number of rectangle primitives $N$}
The ablation study of $N$ is done on Faster R-CNN on a classification attack task in COCO and on \emph{vanishing a car ahead} attack task in the $\text{D}^2$-City dataset.
We investigate the attack performance and the time consumption when we set $N$ to 5, 10 and 15. The results on two datasets are shown in Table~\ref{tab:N_compare_coco} and Table~\ref{tab:N_compare_didi}, respectively. The time consumption is tested on a Nvidia TITAN Xp graphics card.
We find that the performance improvement from increasing $N$ is small when $N$ is large. However, as $N$ increases, it takes more time to generate an adversarial example. A similar phenomenon also can be found on other object detectors.
Considering the balance of efficiency and performance, we set our default $N$ as 10 in all experiments.

\begin{table}[] 
\scriptsize
\centering
\caption{
Different number of rectangle primitives; comparing results on COCO.}
\label{tab:N_compare_coco}
\setlength{\tabcolsep}{1.7mm}{
\begin{tabular}{c|ccccc}
\hline
 Attack Method                 & \makecell[c]{ASR~(\%) \\ (Area Thr~(\%)) }     & \makecell[c]{MAR\\(\%)}  & \makecell[c]{LPIPS \\ ($\times 10^{-3}$) } & \makecell[c]{ADA\\(\%)} & \makecell[c]{Time\\(sec/image)} \\ \hline
LDAP~($N$=5)
& 22.42~(10)  & 15.8 &  16.16 & 76.6 & \textbf{361.26} \\ 
LDAP
& 63.61~(10)   &   9.36    & 12.23    & 73.1 & 369.64 \\ 
LDAP~($N$=15)        
& \textbf{68.50}~(10) & \textbf{9.31} &  \textbf{11.10}  & \textbf{72.2} & 415.34 \\  \hline
\end{tabular}
}
\end{table}

\begin{table}[] 
\scriptsize
\centering
\caption{
Different number of rectangle primitives; comparing results on $\text{D}^2$-City.}
\label{tab:N_compare_didi}
\setlength{\tabcolsep}{1.7mm}{
\begin{tabular}{c|ccccc}
\hline
 Attack Method                 & \makecell[c]{ASR~(\%) \\ (Area Thr~(\%)) }     & \makecell[c]{MAR\\(\%)}  & \makecell[c]{LPIPS \\ ($\times 10^{-3}$) } & \makecell[c]{ADA\\(\%)} & \makecell[c]{Time\\(sec/image)} \\ \hline
LDAP~($N$=5)
& 39.23~(5)   & 6.34      & 53.99    & 82.8 & \textbf{346.47} \\ 
LDAP
& 63.50~(5)   & 4.67      & 49.81   & 79.6 &  367.49 \\ 
LDAP~($N$=15)       
& \textbf{70.80}~(5) & \textbf{4.21} & \textbf{48.97} & \textbf{79.0} & 389.52 \\  \hline
\end{tabular}
}
\end{table}

\subsubsection{Soft-Attack Strategy}
We also evaluate the effectiveness of the soft-attack strategy in the loss function of our LDAP.
The experiments are conducted on MRCN and YOLOv3 on the classification attack task in COCO and \emph{vanishing a car ahead} attack task in the $\text{D}^2$-City dataset.
We compare original LDAP with two other LDAP methods that adopt different attack strategies:
\begin{itemize}
    \item \textbf{LDAP-Attack-1-bbox}~(LDAP-A1) adopts the strategy that only attacks the bounding box with maximum confidence.
    
    \item \textbf{LDAP-Attack-bbox-Equally}~(LDAP-AE) adopts the strategy that equally attacks the bounding boxes that can be detected as the target object.
\end{itemize}
The comparison results are shown in Table~\ref{tab:strategy_compare_coco} and Table~\ref{tab:strategy_compare_didi}. We find that: \textbf{(i)}~only attacking the bounding box with maximum confidence performs poorly for our LDAP, resulting in large MAR and ADA; it is because the bounding box with maximum confidence is continuously changed during the attack, which makes the optimization of primitives unstable, leading to inferior position of patches;
\textbf{(ii)}~the performance of an equally attacking strategy is close to that of our LDAP, but our soft-attack strategy still outperforms this one; it is because most of the bounding boxes considered in this strategy are not accurate, slightly misleading the optimization of primitives and leading to sub-optimal positions of patches.

\begin{table}[] 
\scriptsize
\centering
\caption{
Different attack strategies; comparing results on COCO.}
\label{tab:strategy_compare_coco}
\setlength{\tabcolsep}{1.7mm}{
\begin{tabular}{c|c|cccc}
\hline
Detector       & Attack Method                 & \makecell[c]{ASR~(\%) \\ (Area Thr~(\%)) }     & \makecell[c]{MAR\\(\%)}  & \makecell[c]{LPIPS \\ ($\times 10^{-3}$) } & \makecell[c]{ADA\\(\%)} \\ \hline
\multirow{3}{*}{MRCN}   
& LDAP-A1
& 9.55~(10)   & 19.99       & 21.36    & 81.93    \\ 
& LDAP-AE
& 30.81~(10)   & 14.49      & 15.72   & 72.0     \\ 
& LDAP         
& \textbf{60.93}~(10) & \textbf{9.70} &  \textbf{15.12}   & \textbf{71.1}   \\  \hline
\multirow{3}{*}{YOLOv3} 
& LDAP-A1
& 15.40~(6)  & 12.99       & 12.62    & 73.80    \\ 
& LDAP-AE
& 60.24~(6)   & 6.59      & 9.80   & 66.6     \\ 
& LDAP         
& \textbf{63.75}~(6) & \textbf{5.61}   & \textbf{8.98} & \textbf{62.5} \\ \hline
\end{tabular}
}
\end{table}

\begin{table}[] 
\scriptsize
\centering
\caption{
Different attack strategies; comparing results on $\text{D}^2$-City.}
\label{tab:strategy_compare_didi}
\setlength{\tabcolsep}{1.7mm}{
\begin{tabular}{c|c|cccc}
\hline
Detector       & Attack Method                 & \makecell[c]{ASR~(\%) \\ (Area Thr~(\%)) }     & \makecell[c]{MAR\\(\%)}  & \makecell[c]{LPIPS \\ ($\times 10^{-3}$) } & \makecell[c]{ADA\\(\%)} \\ \hline
\multirow{3}{*}{MRCN}   
& LDAP-A1
& 11.04~(5)   & 17.44      & 55.97   & 82.8   \\ 
& LDAP-AE
& 59.67~(5)   & 4.94      & 51.24   & 79.2    \\ 
& LDAP         
& \textbf{62.22}~(5)  & \textbf{4.68}      & \textbf{49.83}   & \textbf{79.0}     \\  \hline
\multirow{3}{*}{YOLOv3} 
& LDAP-A1
& 57.11~(3)    & 3.30     & 6.91   &  72.3  \\ 
& LDAP-AE
& 62.13~(3)    & 2.97     & 6.85   &  69.2   \\ 
& LDAP         
& \textbf{63.50}~(3)  & \textbf{2.81}     & \textbf{6.70}     & \textbf{65.1}     \\ \hline
\end{tabular}
}
\end{table}

\subsubsection{Region search and texture constraint}
To validate the benefits of both a region search and texture constraint, we set the following two comparison methods: 
\begin{itemize}
    \item \textbf{LDAP without region search}~(LDAP w/o r): To validate the benefit of a region search, the first comparison method uses a position-fixed patch~(similar to AdvPatch~\cite{thys2019fooling}) and adopts the same attack loss and texture loss as our LDAP, called ``LDAP without region search.''
    
    \item \textbf{LDAP without texture constraint}~(LDAP w/o t): To validate the benefit of a texture constraint, the second comparison method shares the same searched region of LDAP, and optimizes texture without any constraints, called ``LDAP without texture constraint.''
\end{itemize}

\begin{figure}[t!]
\centering
\subfloat[]{
\begin{minipage}[]{.23\linewidth} 
\includegraphics[width=\linewidth]{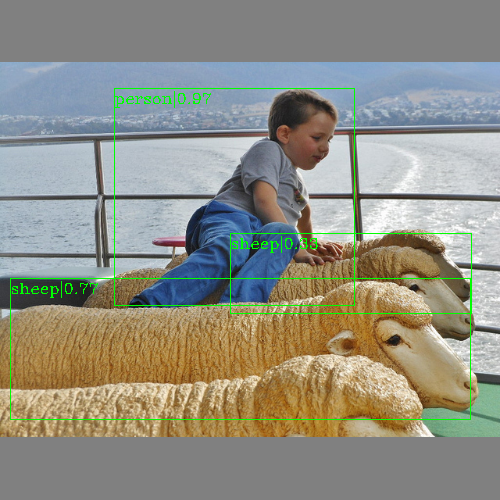}
\\ \vspace{4pt} \quad
\end{minipage}
}
\subfloat[]{
\begin{minipage}[]{.23\linewidth} 
\includegraphics[width=\linewidth]{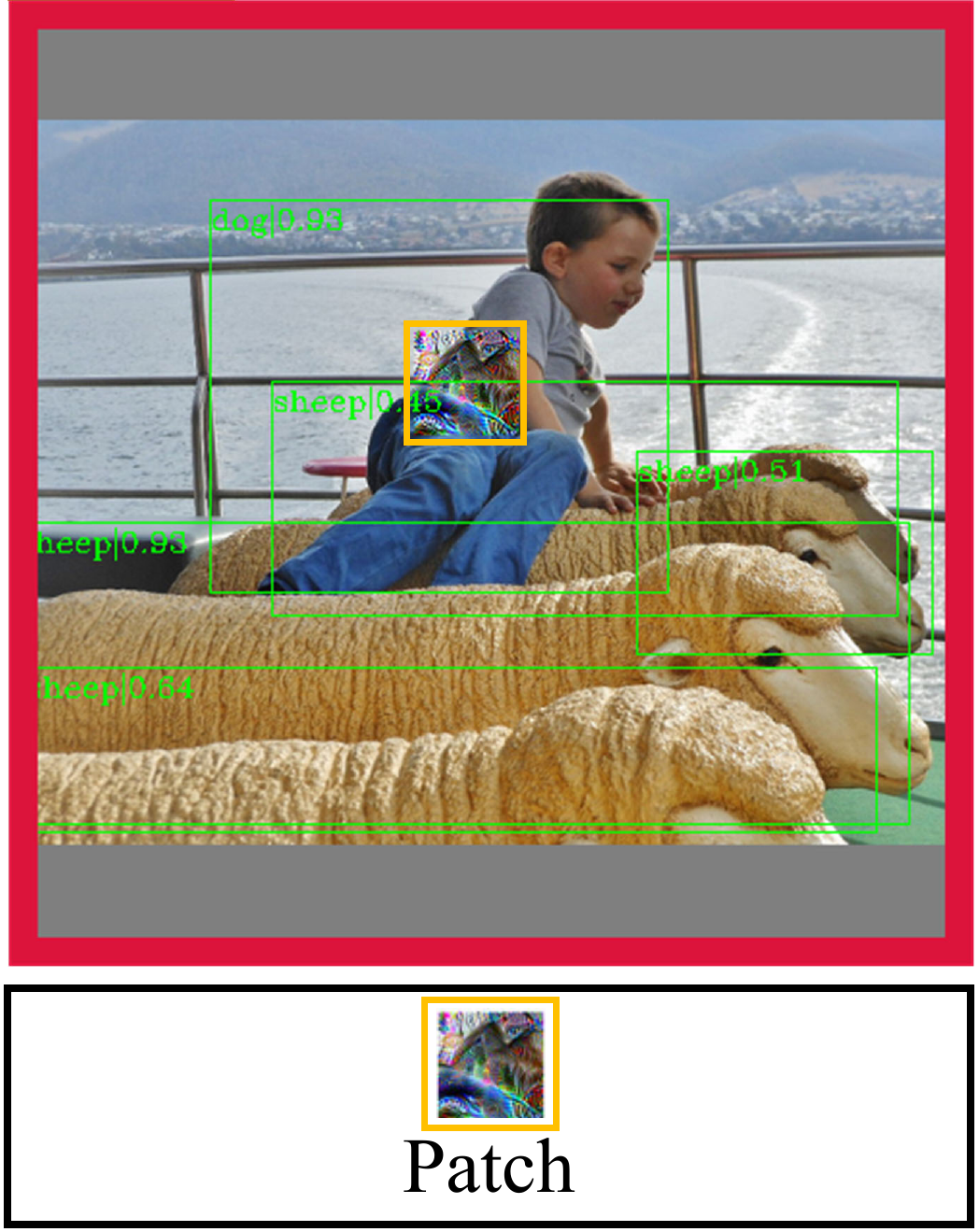}
\end{minipage}
}
\subfloat[]{
\begin{minipage}[]{.23\linewidth} 
\includegraphics[width=\linewidth]{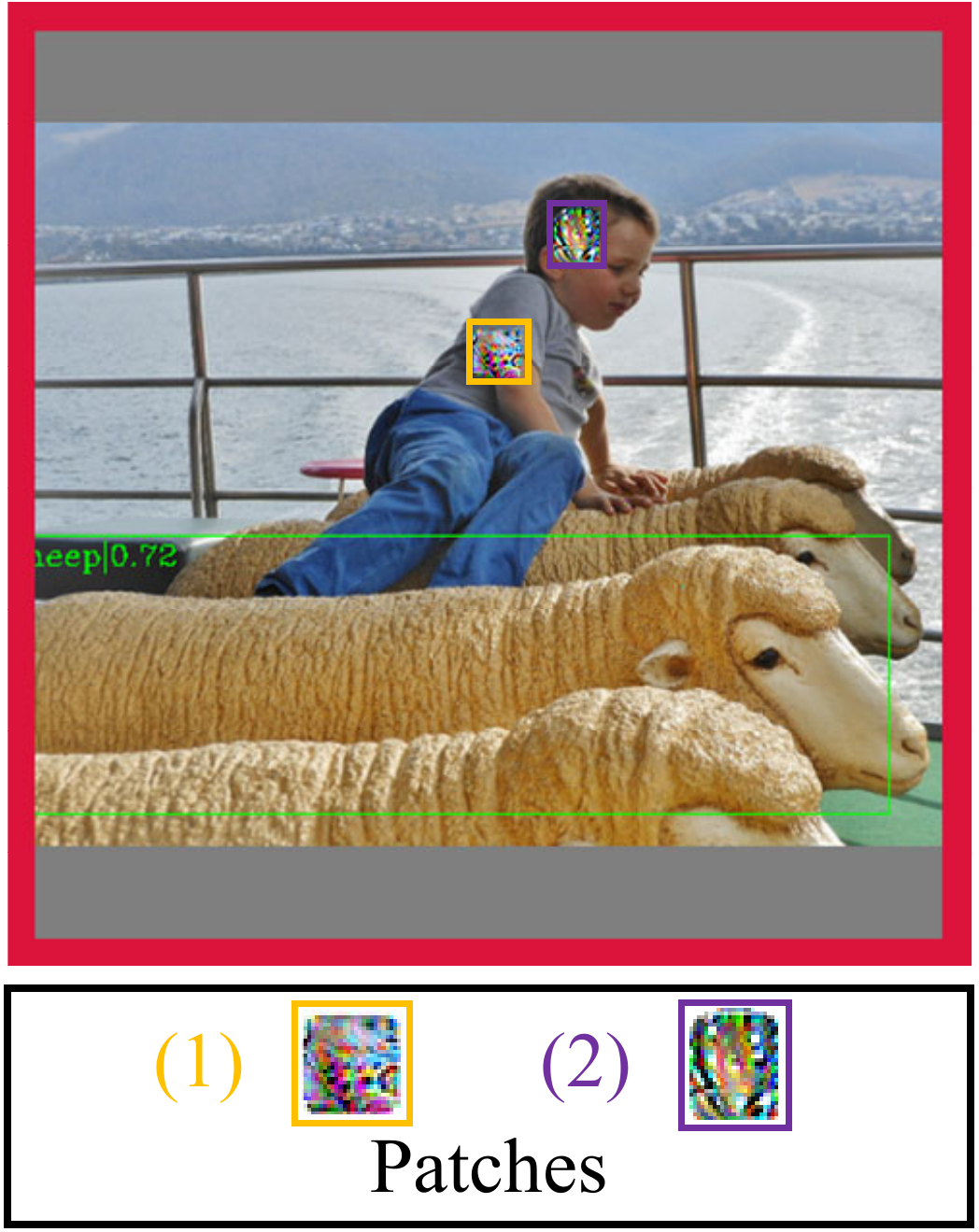}
\end{minipage}
}
\subfloat[]{
\begin{minipage}[]{.23\linewidth} 
\includegraphics[width=\linewidth]{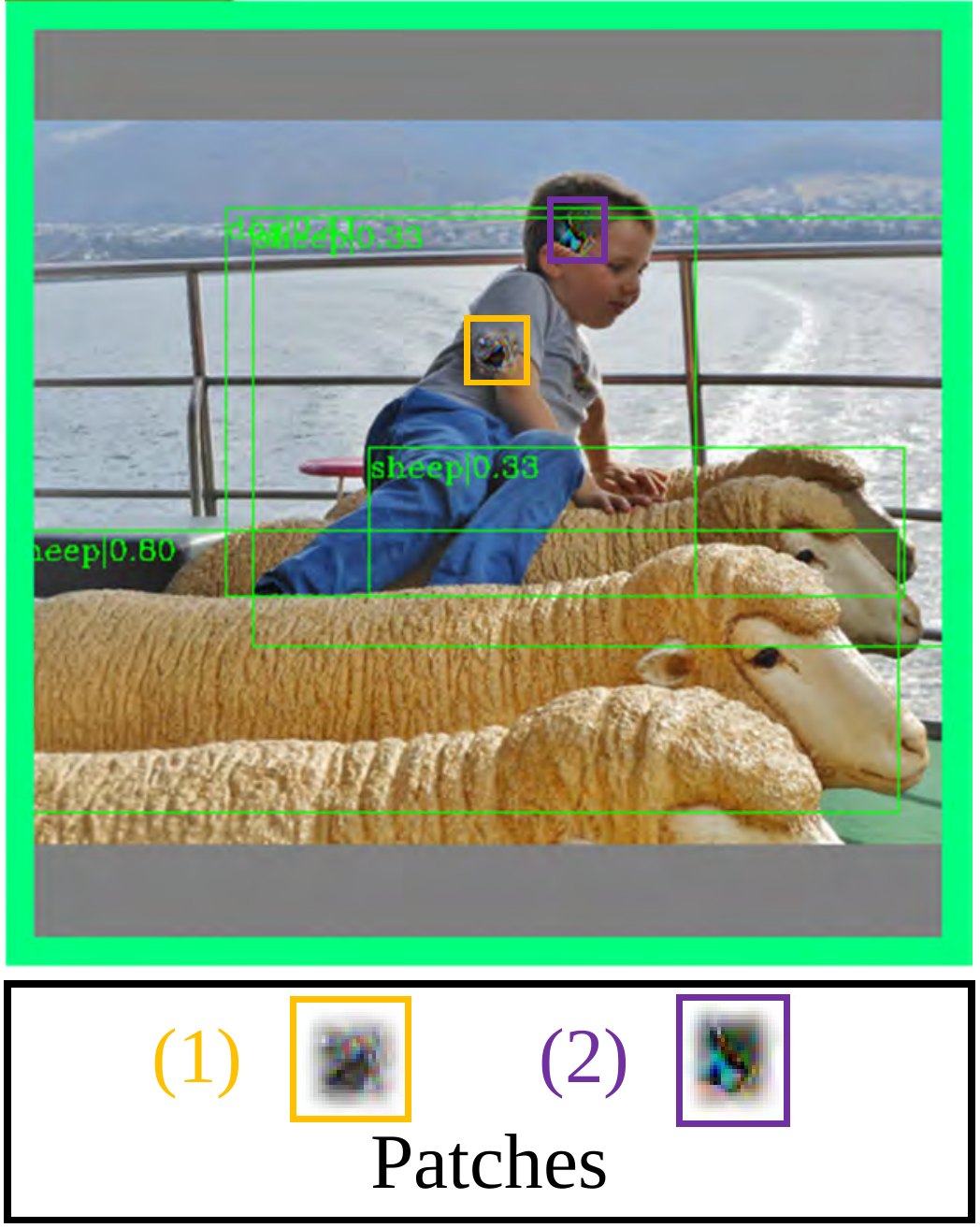}
\end{minipage}
}
\caption{
A comparison of LDAP without region search, LDAP without texture constraint and LDAP on a classification attack on Faster R-CNN on one image from COCO.
The predictions and the probabilities predicted by adversarial patch detectors are:
(a)~original image, person=97\%;
(b)~LDAP w/o r, dog = 93\%, $\text{P}_a$=76.42\%;
(c)~LDAP w/o t, nothing detected, $\text{P}_a$=83.13\%; and
(d)~LDAP, dog = 41\%, $\text{P}_a$=44.73\%.
}
\label{fig:LDAP_rt_compare}
\end{figure}

We compare these two methods with LDAP on a classification attack task in COCO and a \emph{vanishing a car ahead} attack task in the $\text{D}^2$-City dataset.
We show a demonstration of these attacks in COCO in Fig.~\ref{fig:LDAP_rt_compare}. It is obvious that the texture of LDAP is more consistent with the original image and that the patches are smaller.
The comparison results of these methods are shown in Table~\ref{tab:LDAP_rt_compare_coco} and Table~\ref{tab:LDAP_rt_compare_didi}. We find that both the region search and the texture constraint are important for LDAP to achieve a low ADA.

\begin{table}[t!]
\centering
\scriptsize
\caption{
Ablation study on region search and texture constraint on COCO.}
\label{tab:LDAP_rt_compare_coco}
\setlength{\tabcolsep}{1.4mm}{
\begin{tabular}{c|c|cccc}
\hline
Detector       & Attack Method                 & \makecell[c]{ASR~(\%) \\ (Area Thr~(\%)) }     & \makecell[c]{MAR\\(\%)}  & \makecell[c]{LPIPS \\ ($\times 10^{-3}$) } & \makecell[c]{ADA\\(\%)} \\ \hline
\multirow{3}{*}{FRCN} 
& LDAP w/o r
& 10.52~(10)  &  17.70 & 25.42  & 81.8  \\ 
& LDAP w/o t
& \textbf{63.61}~(10)  & \textbf{9.36} & 68.33    &    85.3   \\ 
& LDAP          
& \textbf{63.61}~(10)  & \textbf{9.36}   & \textbf{12.23}   &  \textbf{73.4}   \\ \hline
\multirow{3}{*}{SSD} 
& LDAP w/o r
& 52.95~(7)  &  8.28 &  26.64 & 79.3  \\ 
& LDAP w/o t
& \textbf{63.06}~(7)  &   \textbf{6.39} & 20.26     &    88.8   \\ 
& LDAP          
& \textbf{63.06}~(7)  &   \textbf{6.39} & \textbf{9.33}   &  \textbf{76.6}   \\ \hline
\end{tabular}
}
\end{table}

\begin{table}[t!]
\centering
\scriptsize
\caption{
Ablation study on region search and texture constraint on $\text{D}^2$-City.}
\label{tab:LDAP_rt_compare_didi}
\setlength{\tabcolsep}{1.4mm}{
\begin{tabular}{c|c|cccc}
\hline
Detector       & Attack Method                 & \makecell[c]{ASR~(\%) \\ (Area Thr~(\%)) }     & \makecell[c]{MAR\\(\%)}  & \makecell[c]{LPIPS \\ ($\times 10^{-3}$) } & \makecell[c]{ADA\\(\%)} \\ \hline
\multirow{3}{*}{FRCN} 
& LDAP w/o r
& 28.92~(5)  &   17.92 & 53.74  & 88.2 \\ 
& LDAP w/o t
& \textbf{63.50}~(5)  &  \textbf{4.67} & 62.03 & 93.1  \\   
& LDAP          
& \textbf{63.50}~(5)  &  \textbf{4.67} & \textbf{49.81}  & \textbf{79.6}  \\ \hline
\multirow{3}{*}{SSD} 
& LDAP w/o r
& 55.01~(4)  &  7.45 &  15.02 & 73.2  \\ 
& LDAP w/o t
& \textbf{69.61}~(4)  &   \textbf{3.88}  & 46.51 & 86.6 \\ 
& LDAP
& \textbf{69.61}~(4)  & \textbf{3.88}     & \textbf{5.94}   & \textbf{71.0}    \\ \hline
\end{tabular}
}
\end{table}

\section{Conclusion}
\label{sec:Conclusion}
In this paper, we propose a low-detectable adversarial patch attack method. It aims to reduce the area of patches while simultaneously keeping the texture consistency with the original image when attacking object detectors. That makes our adversarial examples less likely to be recognized by an adversarial patch detector.
We adopt several geometric primitives to model the patches. To enhance attack performance, we adopt a soft-attack strategy. It assigns different weights to bounding boxes that can be detected as a target object.
We prove the effectiveness, low detectability and the threats of our method in both COCO and $\text{D}^2$-City datasets.


{\small
\bibliographystyle{ieee_fullname}
\bibliography{egbib}
}

\end{document}